\documentclass[acmtog,authorversion]{acmart}

\usepackage{booktabs} 

\usepackage[ruled]{algorithm2e} 

\SetAlFnt{\small}
\SetAlCapFnt{\small}
\SetAlCapNameFnt{\small}
\SetAlCapHSkip{0pt}

\copyrightyear{2023}
\acmYear{2023}
\setcopyright{acmlicensed} 
\acmConference[SIGGRAPH '23 Conference Proceedings]{Special Interest Group on Computer Graphics and Interactive Techniques Conference Conference Proceedings}{August 6--10, 2023}{Los Angeles, CA, USA}
\acmBooktitle{Special Interest Group on Computer Graphics and Interactive Techniques Conference Conference Proceedings (SIGGRAPH '23 Conference Proceedings), August 6--10, 2023, Los Angeles, CA, USA}
\acmDOI{10.1145/3588432.3591536}
\acmISBN{979-8-4007-0159-7/23/08}

\usepackage{colortbl}
\usepackage{arydshln}
\usepackage{makecell}
\definecolor{yellow}{rgb}{1,1, 0.7}
\definecolor{lightyellow}{rgb}{1,1, 0.8}
\definecolor{orange}{rgb}{1, 0.85, 0.7}
\definecolor{tablered}{rgb}{1, 0.7, 0.7}
\usepackage[capitalize]{cleveref}
\crefname{section}{Sec.}{Secs.}
\Crefname{section}{Section}{Sections}
\Crefname{table}{Table}{Tables}
\crefname{table}{Tab.}{Tabs.}

\usepackage{amsmath,amsfonts,bm}
\usepackage{multirow}
\usepackage{tabularx}
\usepackage{graphicx}
\usepackage{adjustbox}
\usepackage{xspace}
\usepackage{xfrac}

\newcommand{\modelname}{BakedSDF}

\newcommand{\pos}{\mathbf{x}}
\newcommand{\origin}{\mathbf{o}}
\newcommand{\raydir}{\mathbf{d}}

\newcommand{\density}{\tau}
\newcommand{\col}{\mathbf{c}}

\newcommand{\mset}[1]{\left\{\kern-.5em\left\{ #1 \right\}\kern-.5em\right\}}
\newcommand{\mmset}[1]{\{\kern-.4em\{ #1 \}\kern-.4em\}}

\newcommand{\norm}[1]{\left\Vert#1\right\Vert}

\newcommand{\parr}[1]{\left (#1\right )}

\def \etal{{et al}.\xspace}

\newcommand{\ie}{{i.e.}\xspace}

\def\eqref#1{equation~\ref{#1}}

\def\1{\bm{1}}

\def\rvx{{\mathbf{x}}}

\def\vec1{{\bm{1}}}

\DeclareMathAlphabet{\mathsfit}{\encodingdefault}{\sfdefault}{m}{sl}
\SetMathAlphabet{\mathsfit}{bold}{\encodingdefault}{\sfdefault}{bx}{n}

\def\gL{{\mathcal{L}}}

\def\gS{{\mathcal{S}}}

\newcommand{\E}{\mathbb{E}}

\begin{document}

\title[BakedSDF: Meshing Neural SDFs for Real-Time View Synthesis]{BakedSDF: Meshing Neural SDFs for Real-Time View Synthesis}

\author{Lior Yariv}
\authornote{Both authors contributed equally to this research.}
\authornote{Work done while interning at Google.}
\affiliation{
\institution{Weizmann Institute of Science}
\country{Israel}
}
\affiliation{
\institution{Google Research}
\country{United Kingdom}
}

\author{Peter Hedman}
\authornotemark[1]
\affiliation{
\institution{Google Research}
\country{United Kingdom}
}

\author{Christian Reiser}
\authornotemark[2]
\affiliation{
\institution{Tübingen AI Center}
\country{Germany}
}
\affiliation{
\institution{Google Research}
\country{United Kingdom}
}

\author{Dor Verbin}
\affiliation{
\institution{Google Research}
\country{United States of America}
}

\author{Pratul P. Srinivasan}
\affiliation{
\institution{Google Research}
\country{United States of America}
}

\author{Richard Szeliski}
\affiliation{
\institution{Google Research}
\country{United States of America}
}

\author{Jonathan T. Barron}
\affiliation{
\institution{Google Research}
\country{United States of America}
}

\author{Ben Mildenhall}
\affiliation{
\institution{Google Research}
\country{United States of America}
}

\renewcommand\shortauthors{Lior Yariv*, Peter Hedman*, \etal}

\begin{abstract}
We present a method for reconstructing high-quality meshes of large unbounded real-world scenes suitable for photorealistic novel view synthesis. We first optimize a hybrid neural volume-surface scene representation designed to have well-behaved level sets that correspond to surfaces in the scene. We then bake this representation into a high-quality triangle mesh, which we equip with a simple and fast view-dependent appearance model based on spherical Gaussians. Finally, we optimize this baked representation to best reproduce the captured viewpoints, resulting in a model that can leverage accelerated polygon rasterization pipelines for real-time view synthesis on commodity hardware. Our approach outperforms previous scene representations for real-time rendering in terms of accuracy, speed, and power consumption, and produces high quality meshes that enable applications such as appearance editing and physical simulation.
\end{abstract}

%


%
%


\begin{teaserfigure}
    \centering
    \begin{tabular}{@{}c@{\,\,}c@{}}
    \begin{tabular}{@{}c@{}}
    \small (a) Extracted mesh \\
    \includegraphics[width=0.515\linewidth]{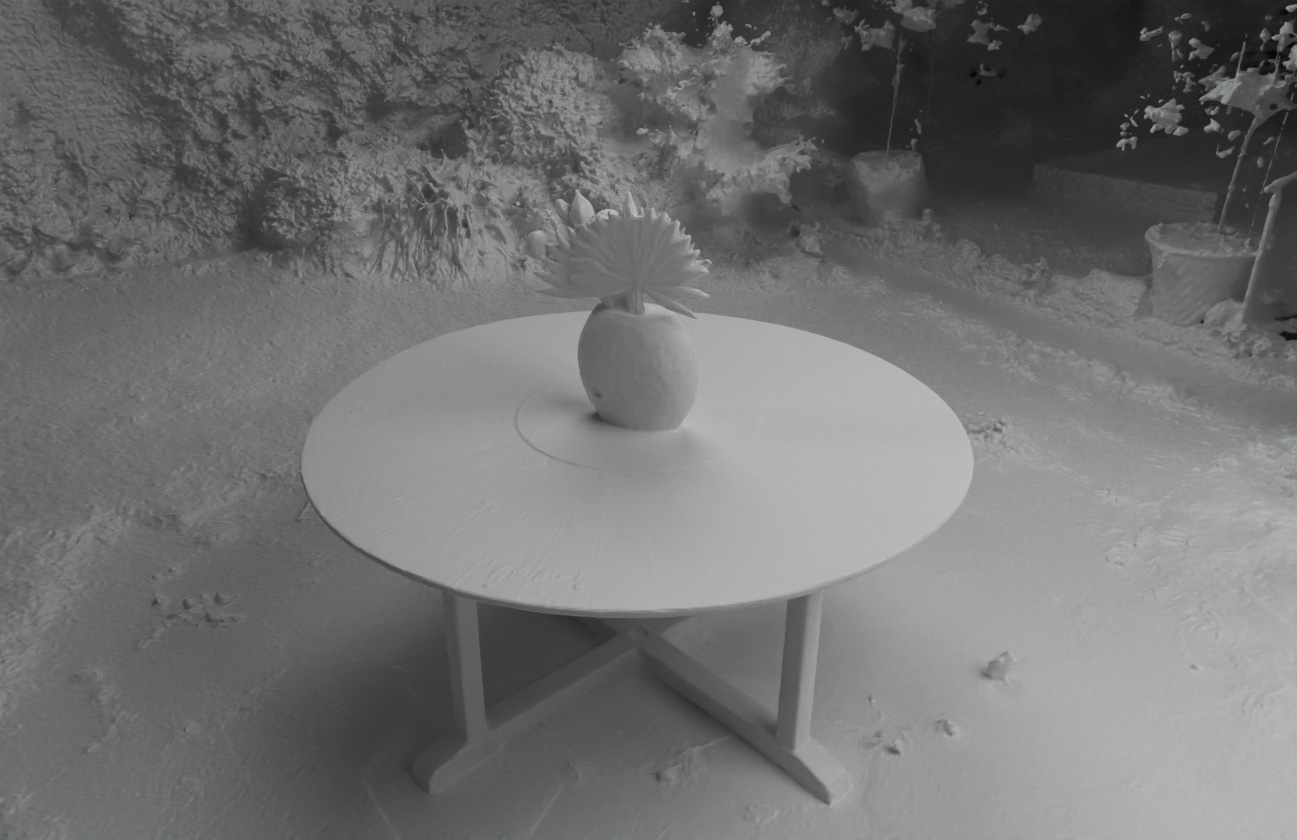}
    \end{tabular}
    \begin{tabular}{@{}c@{\,\,}c@{}}
    \small (b) Rendering ($\sim$105 FPS) & \small (c) Diffuse/specular components \\
    \includegraphics[width=0.235\linewidth]{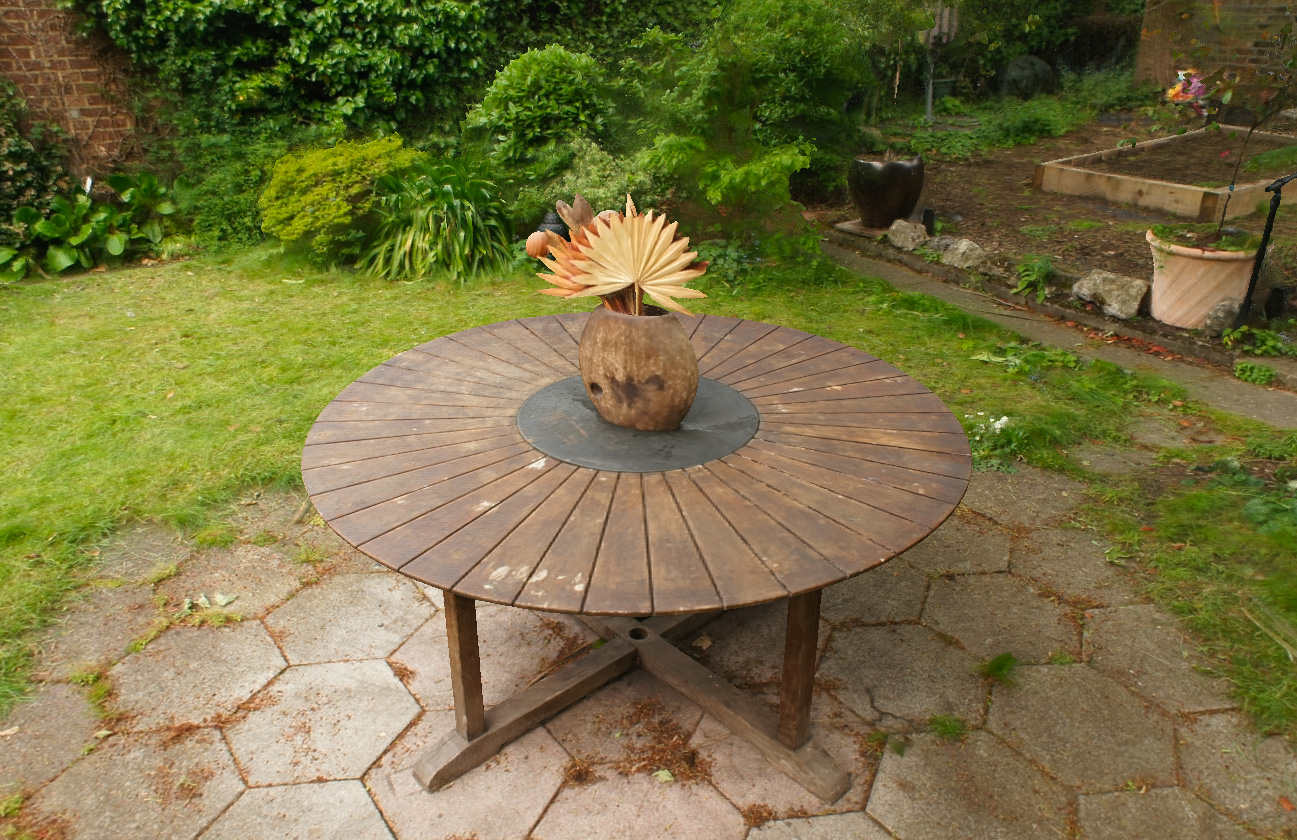} &
    \includegraphics[width=0.235\linewidth]{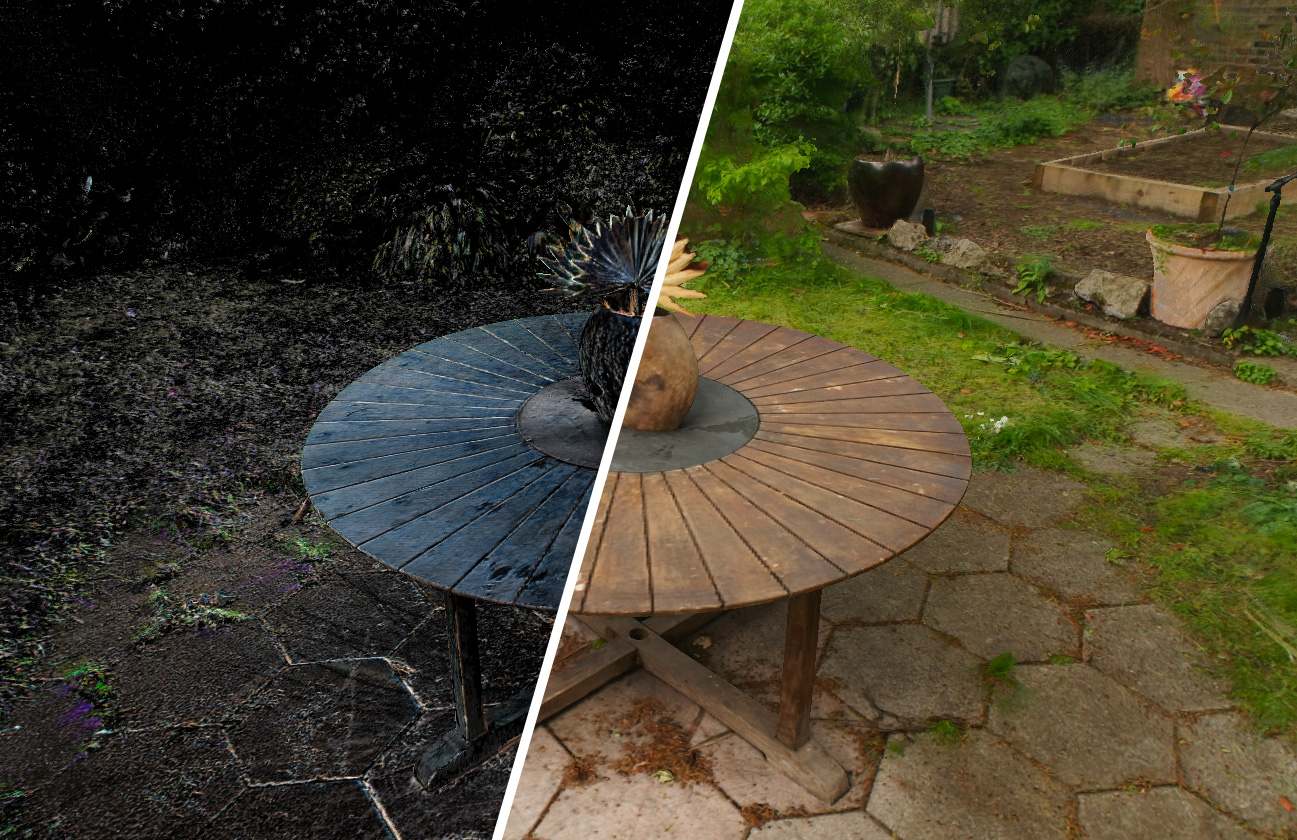} \\
    \small (d) Appearance editing & \small (e) Physics simulation \\
    \includegraphics[width=0.235\linewidth]{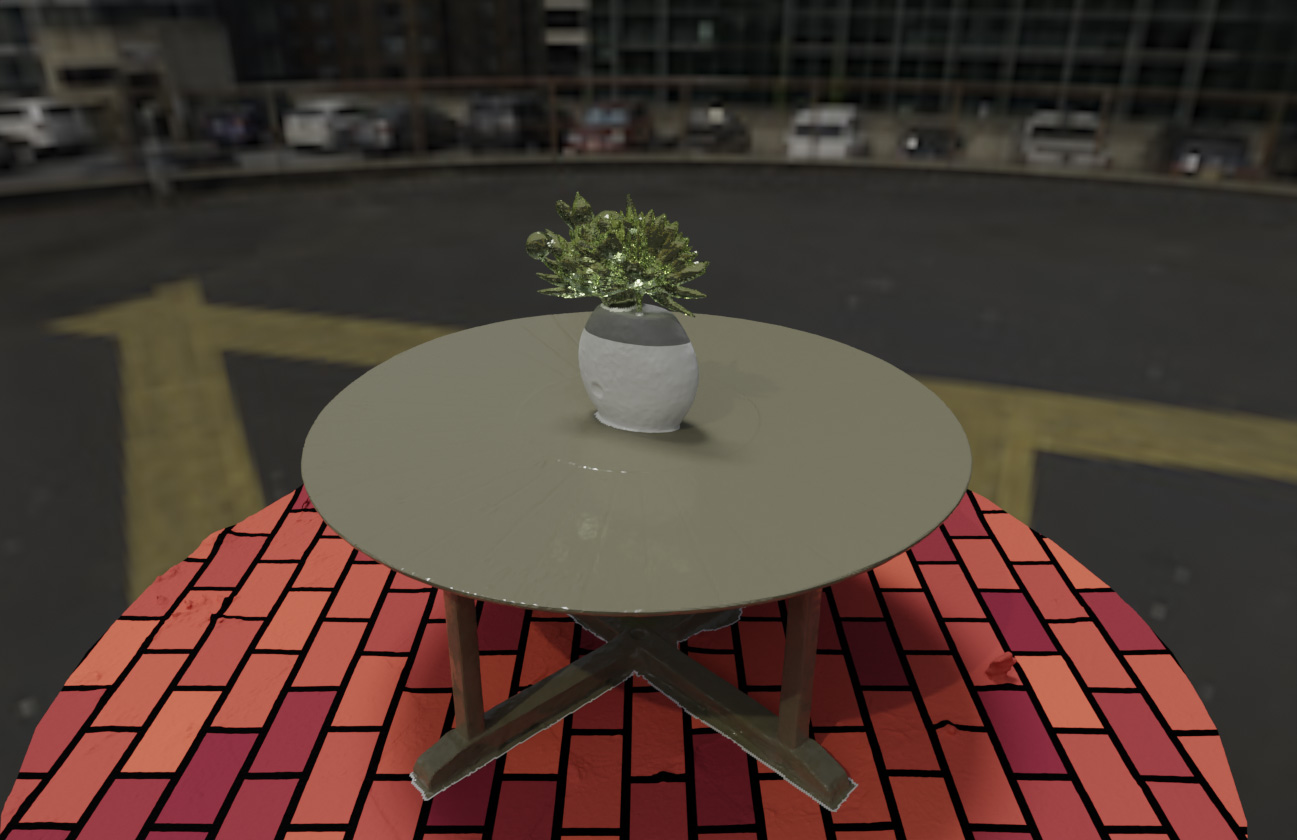} &
    \includegraphics[width=0.235\linewidth]{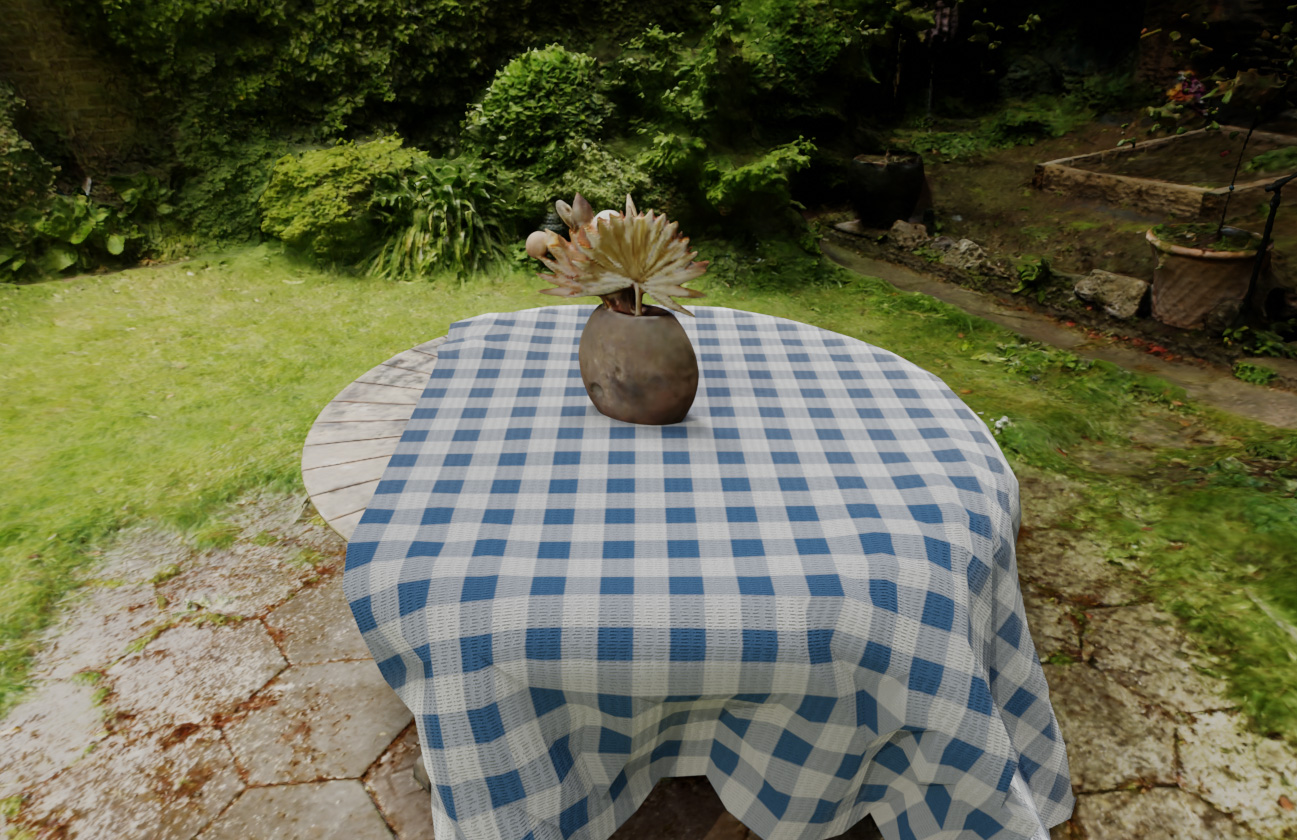} \\
    \end{tabular}
    \end{tabular}
    \vspace{-10pt}
    \caption{Our method, \emph{\modelname}, optimizes a neural surface-volume representation of a complex real-world scenes and (a) ``bakes`` that representation into a high-resolution mesh. These meshes (b) can be rendered in real time on commodity hardware, and support other applications such as (c) separating material components, (d) appearance editing with accurate cast shadows, and (e) physics simulation for inserted objects. aInteractive demo at \url{https://bakedsdf.github.io/}.
    }
    \label{fig:teaser}
    \vspace{5pt}
\end{teaserfigure}


\maketitle


\section{Introduction}
\label{sec:intro}

Current top-performing approaches for novel view synthesis --- the task of using captured images to recover a 3D representation that can be rendered from unobserved viewpoints --- are largely based on Neural Radiance Fields (NeRF)~\cite{mildenhall2020nerf}.
By representing a scene as a continuous volumetric function parameterized by a multilayer perceptron (MLP), NeRF is able to produce photorealistic renderings that exhibit detailed geometry and view-dependent effects. Because the MLP underlying a NeRF is expensive to evaluate and must be queried hundreds of times \emph{per pixel}, rendering a high resolution image from a NeRF is typically slow.

Recent work has improved NeRF rendering performance by trading compute-heavy MLPs for discretized volumetric representations such as voxel grids. However, these approaches require substantial GPU memory and custom volumetric raymarching code and are not amenable to real-time rendering on commodity hardware, since modern graphics hardware and software is oriented towards rendering polygonal surfaces rather than volumetric fields.

While current NeRF-like approaches are able to recover high-quality real-time-renderable meshes of individual objects with simple geometry~\cite{boss2022-samurai}, reconstructing detailed and well-behaved meshes from captures of real-world unbounded scenes (such as the ``360 degree captures'' of \citet{barron2022mipnerf360}) has proven to be more difficult. 
Recently, MobileNeRF~\cite{chen2022mobilenerf} addressed this problem by training a NeRF whose volumetric content is restricted to lie on the faces of a polygon mesh, then baking that NeRF into a texture map.
Though this approach yields reasonable image quality, MobileNeRF initializes the scene geometry as a collection of axis-aligned tiles that turns into a textured polygon ``soup'' after optimization. The resulting geometry is less suitable for common graphics applications such as texture editing, relighting, and physical simulation.



In this work, we demonstrate how to extract high-quality meshes from a NeRF-like neural volumetric representation.
Our system, which we call \modelname, extends the hybrid volume-surface neural representation of VolSDF~\cite{yariv2021volume} to represent unbounded real-world scenes.
This representation is designed to have a well-behaved zero level set corresponding to surfaces in the scene, which lets us extract high-resolution triangle meshes using marching cubes.
%

Our key idea is to define the SDF in \emph{contracted coordinate space}~\cite{barron2022mipnerf360}, as it has these advantages: It more strongly regularizes distant content, and it allows us to also extract the mesh in contracted space which distributes the triangle budget better (more in the center, fewer in the periphery).

We then equip this mesh with a fast and efficient view-dependent appearance model based on spherical Gaussians, which is fine-tuned to reproduce the input images of the scene.
The output of our system can be rendered at real-time frame rates on commodity devices, and 
we show that our real-time rendering system outperforms prior work in terms of realism, speed, and power consumption.
Additionally we show that (unlike comparable prior work) the mesh produced by our model is accurate and detailed, enabling standard graphics applications such as appearance editing and physics simulation.

To summarize, our key contributions are: \vspace{5pt}
\begin{enumerate}
\item High-quality neural surface reconstruction of unbounded real-world scenes,
\item a framework for real-time rendering of these scenes in a browser, and
\item we demonstrate that spherical Gaussians are a practical representation of view-dependence appearance for view-synthesis.
\end{enumerate}

\newpage

\section{Related Work}

View synthesis, \ie, the task of rendering novel views of a scene given a set of captured images, is a longstanding problem in the fields of computer vision and graphics. In scenarios where the observed viewpoints are sampled densely, synthesizing new views can be done with light field rendering --- straightforward interpolation into the set of observed rays~\cite{gortler1996lumigraph,levoy1996light}. However, in practical settings where observed viewpoints are captured more sparsely, reconstructing a 3D representation of the scene is crucial for rendering convincing novel views. Most classical approaches for view synthesis use triangle meshes (typically reconstructed using a pipeline consisting of multi-view stereo~\cite{Furukawa_Fnt15,schoenberger2016mvs}, Poisson surface reconstruction~\cite{SGP:SGP06:061-070,kazhdan2013screened}, and marching cubes~\cite{lorensen1987marching}) as the underlying 3D scene representation, and render novel views by reprojecting observed images into each novel viewpoint and blending them together using either heuristically-defined~\cite{debevec:1996:vdtm,buehler2001unstructured,wood:2000:slf} or learned~\cite{hedman2018deep,riegler2020free,riegler2021stable} blending weights. Although mesh-based representations are well-suited for real-time rendering with accelerated graphics pipelines,
the meshes produced by these approaches tend to have inaccurate geometry in regions with fine details or complex materials, which leads to errors in rendered novel views. Alternatively, point-based representations~\cite{ruckert2021adop,kopanas2021point} are better suited for modeling thin geometry, but cannot be rendered efficiently without visible cracks or unstable results when the camera moves.

Most recent approaches to view synthesis sidestep the difficulty of high-quality mesh reconstruction by using volumetric representations of geometry and appearance, such as voxel grids~\cite{szeliski1999planesweep,vogiatzis2007volumetric,Soft3DReconstruction,lombardi2019neuralvolumes} or multiplane images~\cite{srinivasan2019pushing,Zhou_StereoMagn_SG18,wizadwongsa2021nex}. These representations are well-suited to gradient-based optimization of a rendering loss, so they can be effectively optimized to reconstruct detailed geometry seen in the input images. The most successful of these volumetric approaches is Neural Radiance Fields (NeRF)~\cite{mildenhall2020nerf}, which forms the basis for many state-of-the-art view synthesis methods (see  \citet{tewari2022advances} for a review).
NeRF represents a scene as a continuous volumetric field of matter that emits and absorbs light, and renders an image using volumetric ray-tracing. NeRF uses an MLP to map from a spatial coordinate to a volumetric density and emitted radiance, and that MLP must be evaluated at a set of sampled coordinates along a ray to yield a final color.









Subsequent works have proposed modifying NeRF's representation of scene geometry and appearance for improved quality and editability. Ref-NeRF~\cite{verbin2022ref} reparameterizes NeRF's view-dependent appearance to enable appearance editing and improve the reconstruction and rendering of specular materials. Other works~\cite{boss2021nerd,kuang2022neroic,srinivasan2021nerv,zhang2021physg,zhang2021nerfactor} attempt to decompose a scene's view-dependent appearance into material and lighting properties. In addition to modifying NeRF's representation of appearance, papers including UNISURF~\cite{oechsle2021unisurf}, VolSDF~\cite{yariv2021volume}, NeuS~\cite{wang2021neus}, MetaNLR++~\cite{bergman2021metanlr}, and NeuMesh~\cite{neumesh} augment NeRF's fully-volumetric representation with hybrid volume-surface models, but do not target real-time rendering and show results only for objects and bounded scenes.

The MLP NeRF uses to represent a scene is usually large and expensive to evaluate, and this means that a NeRF is slow to train (hours or days per scene) and slow to render (seconds or minutes per megapixel). While rendering can be accelerated with a sampling network that reduces the MLP queries per ray~\cite{neff2021donerf},
recent methods have improved both training and render time by replacing the large MLP with a voxel grid~\cite{karnewar2022relu,sun2022direct}, a grid of small MLPs~\cite{reiser2021kilonerf}, low-rank~\cite{chen2022tensorf} or sparse~\cite{yu2021plenoxels} grid representations, or a multiscale hash encoding with a small MLP~\cite{muller2022instant}.

While these representations reduce the computation required for both training and rendering (at the cost of increased storage), rendering can be further accelerated by precomputing and storing, \ie, ``baking'', a trained NeRF into a more efficient representation. SNeRG~\cite{hedman2021snerg}, FastNeRF~\cite{garbin2021fastnerf}, Plenoctrees~\cite{yu2021plenoctrees}, and Scalable Neural Indoor Scene Rendering~\cite{wu2022snisr} all bake trained NeRFs into sparse volumetric structures and use simplified models of view-dependent appearance to avoid evaluating an MLP at each sample along each ray. These methods have enabled real-time rendering of NeRFs on high-end hardware, but their use of volumetric raymarching precludes real-time performance on commodity hardware.
Concurrent to our work, \citet{reiser2023merf} developed Memory-Efficient Radiance Fields (MERF), a compressed representation volumetric for unbounded scenes that facilitates fast rendering on commodity hardware. When compared with our meshes, this volumetric representation achieves higher quality scores, but requires more memory, needs a complex renderer, and is not straightforward to use for downstream graphics applications such as physics simulation. Please refer to the MERF paper for a direct comparison with our method.

\section{Preliminaries}

In this section, we describe the neural volumetric representation that NeRF~\cite{mildenhall2020nerf} uses for view synthesis as well as improvements introduced by mip-NeRF 360~\cite{barron2022mipnerf360} for representing unbounded ``360 degree'' scenes.



A NeRF is a 3D scene representation consisting of a learned function that maps a position $\pos$ and outgoing ray direction $\raydir$ to a volumetric density $\density$ and color $\col$. To render the color of a single pixel in a target camera view, we first compute the ray corresponding to that pixel $\mathbf r = \origin + t \raydir$, and then evaluate the NeRF at a series of points $\{t_i\}$ along the ray. The resulting outputs $\density_i, \col_i$ at each point are composited together into a single output color value $\mathbf C$:
\begin{equation}
    \mathbf C = \sum_{i} \exp\left(-\sum_{j < i} \density_j \delta_j \right) \left(1 - \exp\left(-\density_i \delta_i \right)\right) \col_i\,, \quad \delta_i = t_i - t_{i-1} \,. \label{eq:rendering}
\end{equation}
This definition of $\mathbf C$ is a quadrature-based approximation of the volume rendering equation~\cite{max1995optical}.

NeRF parametrizes this learned function using an MLP whose weights are optimized to implicitly encode the geometry and color of the scene: A set of training input images and their camera poses are converted into a set of (ray, color) pairs, and gradient descent is used to optimize the MLP weights such that the rendering of each ray resembles its corresponding input color. Formally, NeRF minimizes a loss between the ground truth color $\mathbf{C}_{\mathrm{gt}}$ and the color $\mathbf{C}$ produced in Equation~\ref{eq:rendering}, averaged over all training rays:
\begin{equation}
\label{eqn:loss}
    \mathcal{L_{\mathrm{data}}} = \mathbb{E}\left[ \norm{\mathbf{C} - \mathbf{C}_{\mathrm{gt}}}^2\right] \, .
\end{equation}
If the input images provide sufficient coverage of the scene (in terms of multiview 3D constraints), this simple process yields a set of MLP weights that accurately describe the scene's 3D volumetric density and appearance.

Mip-NeRF 360~\cite{barron2022mipnerf360} extends the basic NeRF formulation to reconstruct and render real-world ``360 degree'' scenes where cameras can observe unbounded scene content in all directions. Two improvements introduced in mip-NeRF 360 are the use of a contraction function and a proposal MLP. The contraction function maps unbounded scene points in $\mathbb R^3$ to a bounded domain:
\begin{equation}
\operatorname{contract}(\mathbf{x})  = \begin{cases}
\mathbf{x} & \norm{\mathbf{x}} \leq 1\\
\left(2  - \frac{1}{\norm{\mathbf{x}}}\right)\frac{\mathbf{x}}{\norm{\mathbf{x}}} & \norm{\mathbf{x}} > 1
\end{cases}  \, ,
\label{eq:contract}
\end{equation}
which produces contracted coordinates that are well-suited to be positionally encoded as inputs to the MLP. Additionally, mip-NeRF 360 showed that large unbounded scenes with detailed geometry require prohibitively large MLPs and many more samples along each ray than is tractable in the original NeRF framework. Mip-NeRF 360 therefore introduced a proposal MLP: a much smaller MLP that is trained to bound the density of the actual NeRF MLP. This proposal MLP is used in a hierarchical sampling procedure that efficiently generates a set of input samples for the NeRF MLP that are tightly focused around non-empty content in the scene.

\begin{figure}[t!]
\centering
    \includegraphics[width=1.0\linewidth]{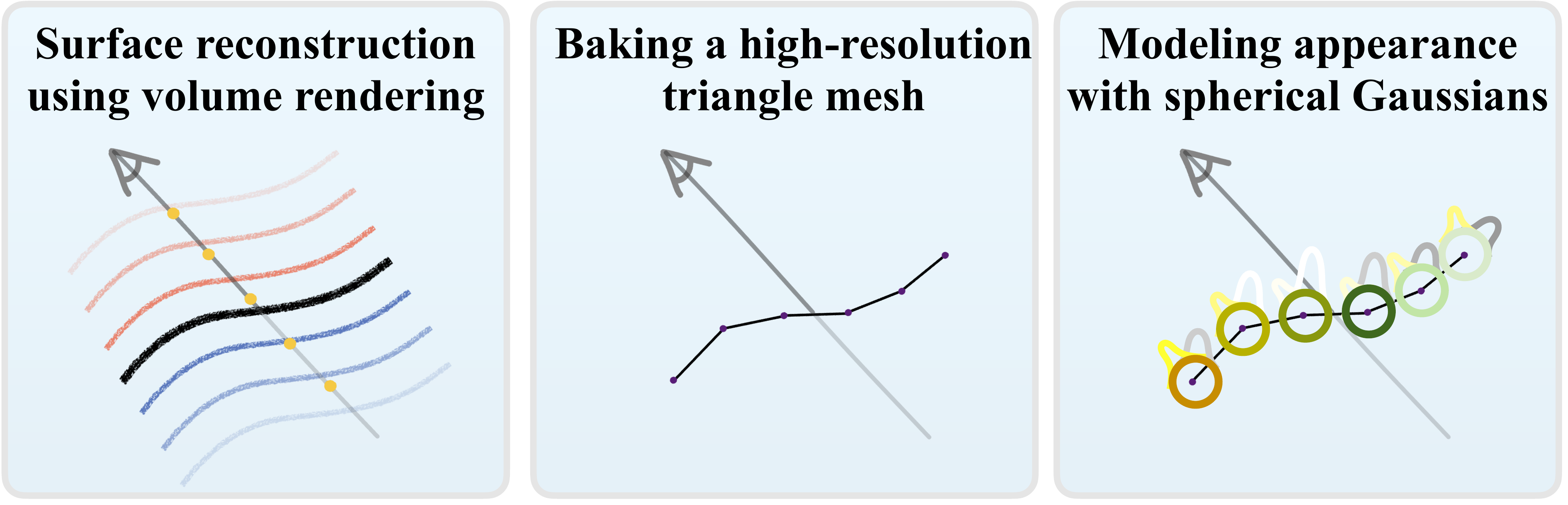}
    \vspace{-20pt}
    \caption{An illustration of the three stages of our method. We first reconstruct the scene using a surface-based volumetric representation (Section~\ref{s:surfacerecon}), then bake it into a high-quality mesh (Section~\ref{s:bakemesh}), and finally optimize a view-dependent appearance model based on spherical Gaussians (Section~\ref{s:viewdep}).
    } \label{fig:pipeline}
    \vspace{-10pt}
\end{figure}

\section{Method}

Our method is composed of three stages, which are visualized in Figure~\ref{fig:pipeline}. First we optimize a surface-based representation of the geometry and appearance of a scene using NeRF-like volume rendering. Then, we ``bake'' that geometry into a mesh, which we show is accurate enough to support convincing appearance editing and physics simulation. Finally, we train a new appearance model that uses spherical Gaussians (SGs) embedded within each vertex of the mesh, which replaces the expensive NeRF-like appearance model from the first step. The resulting 3D representation that results from this approach can be rendered in real-time on commodity devices, as rendering simply requires rasterizing a mesh and querying a small number of spherical Gaussians.

\begin{figure*}[ht!]
\centering
    \begin{tabular}[width=\pagewidth]{@{\hskip0pt}c@{\hskip3pt}c@{\hskip3pt}c@{\hskip3pt}c}
     \includegraphics[width=0.245\linewidth]{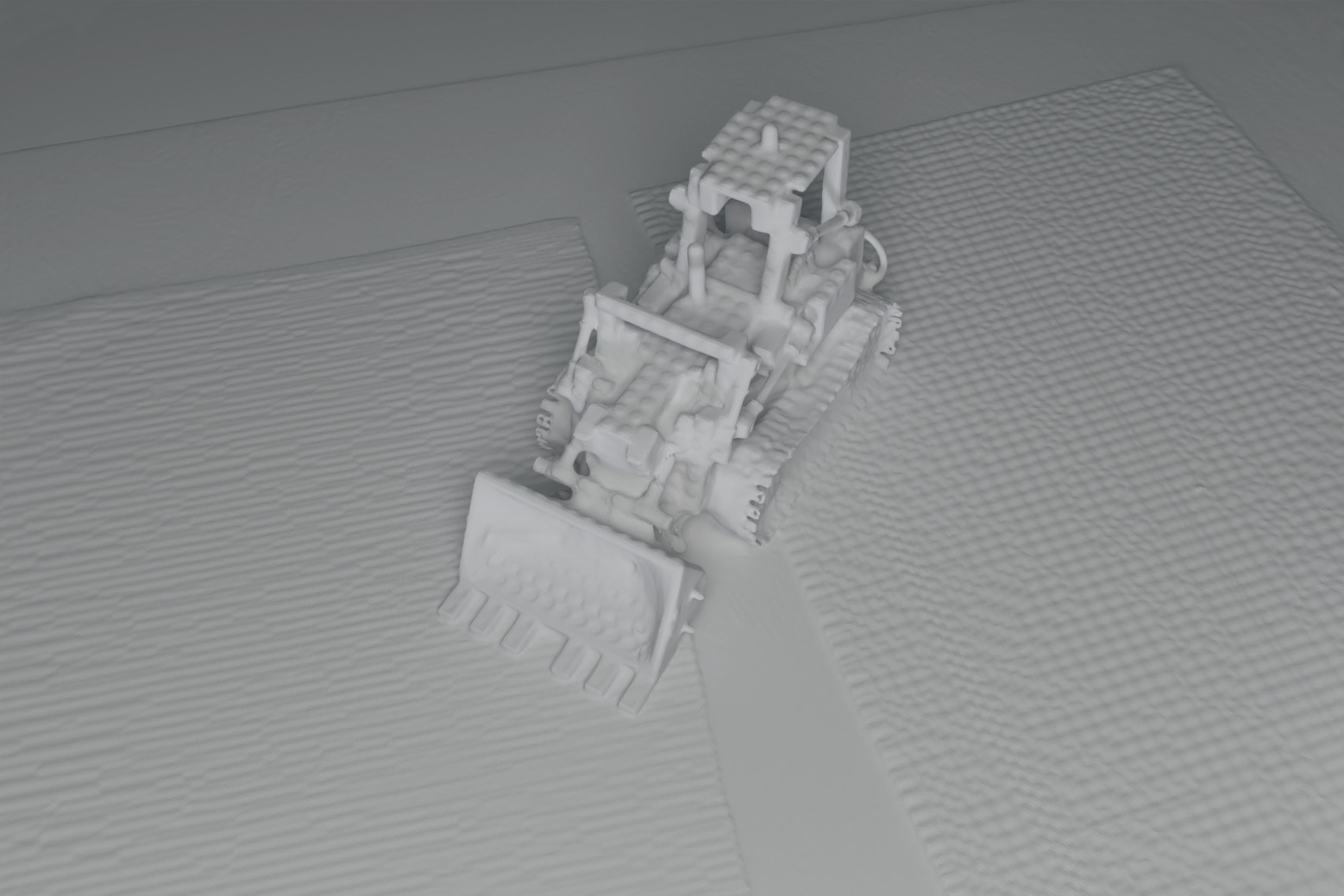} &
     \includegraphics[width=0.245\linewidth]{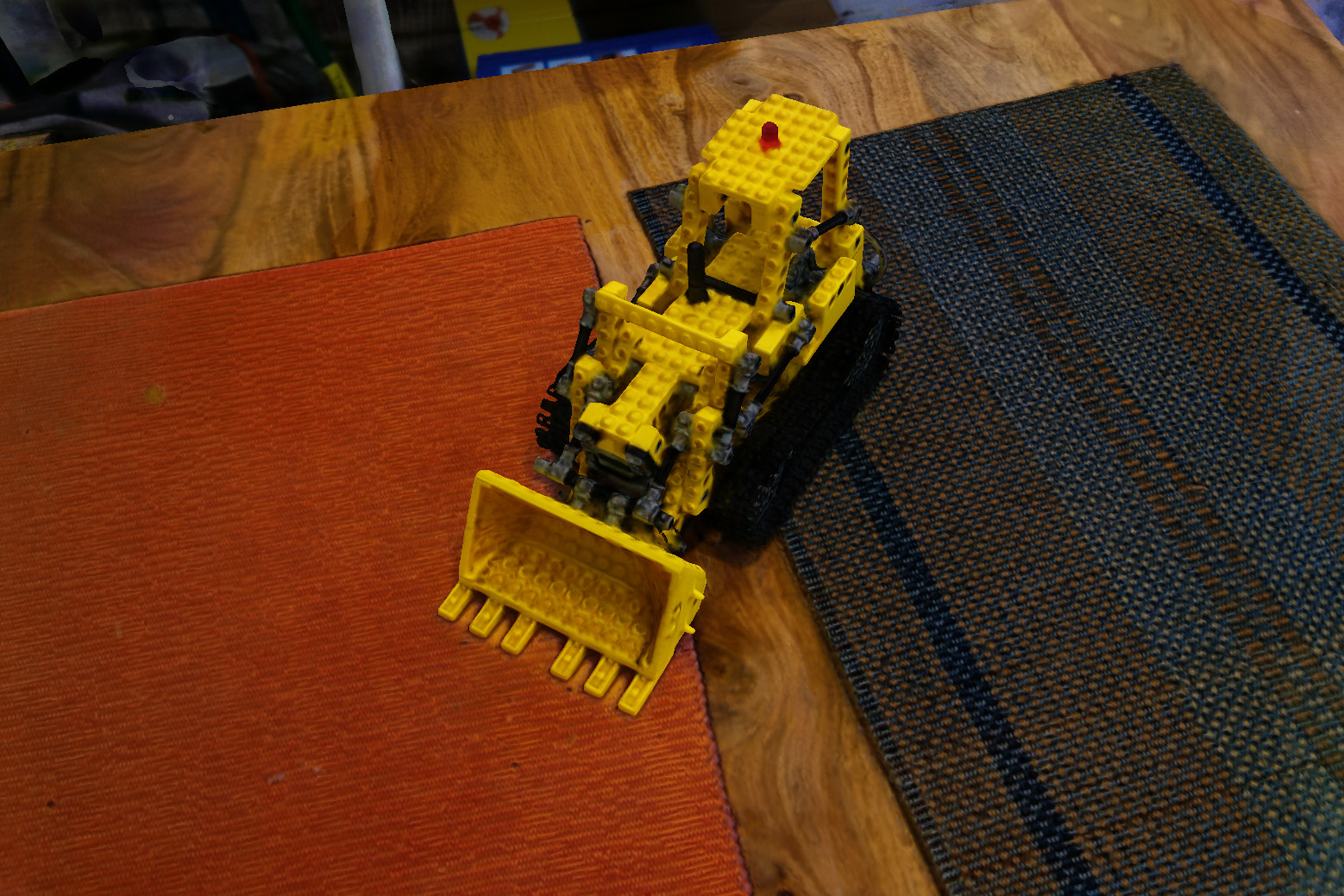} &
     \includegraphics[width=0.245\linewidth]{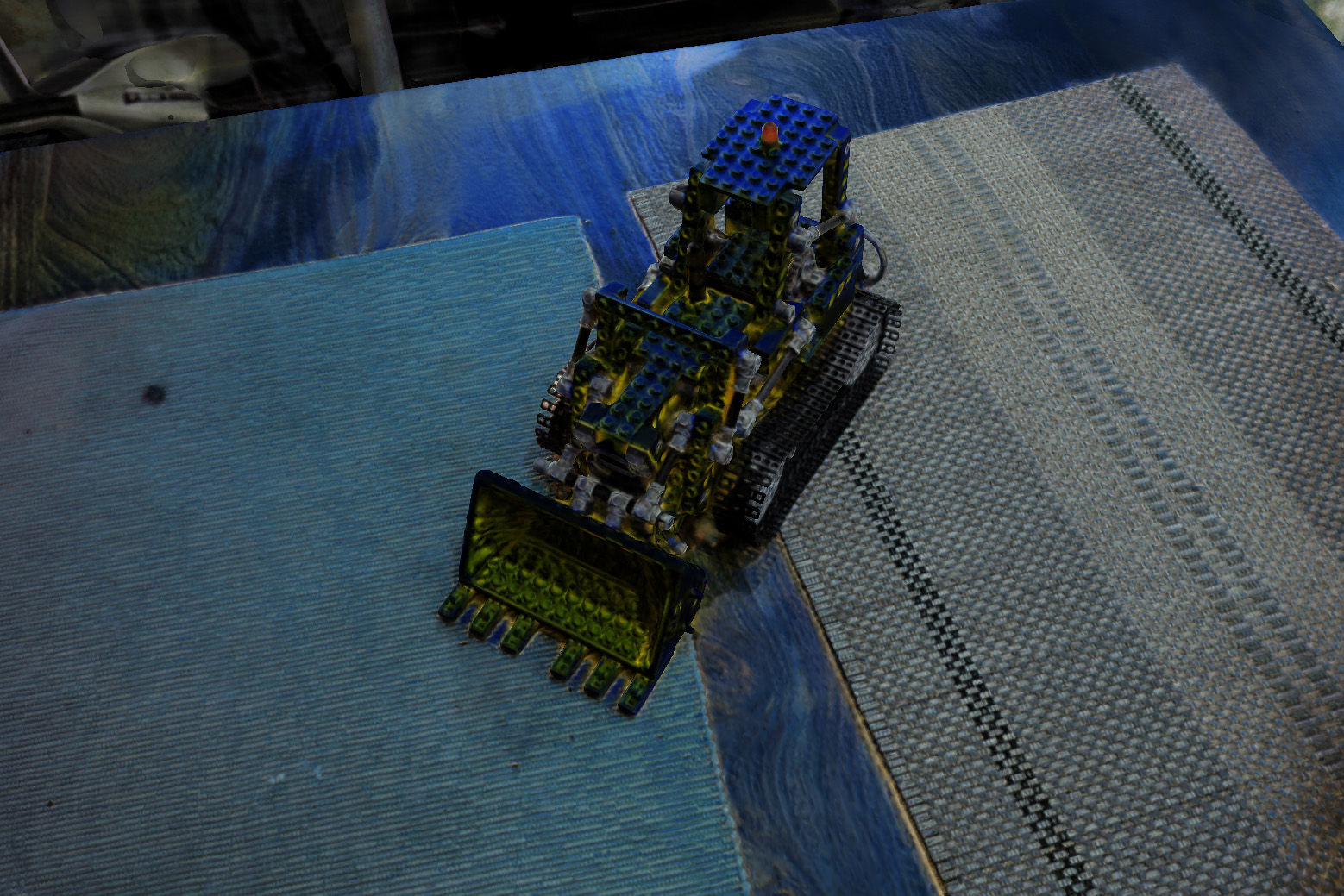} &
     \includegraphics[width=0.245\linewidth]{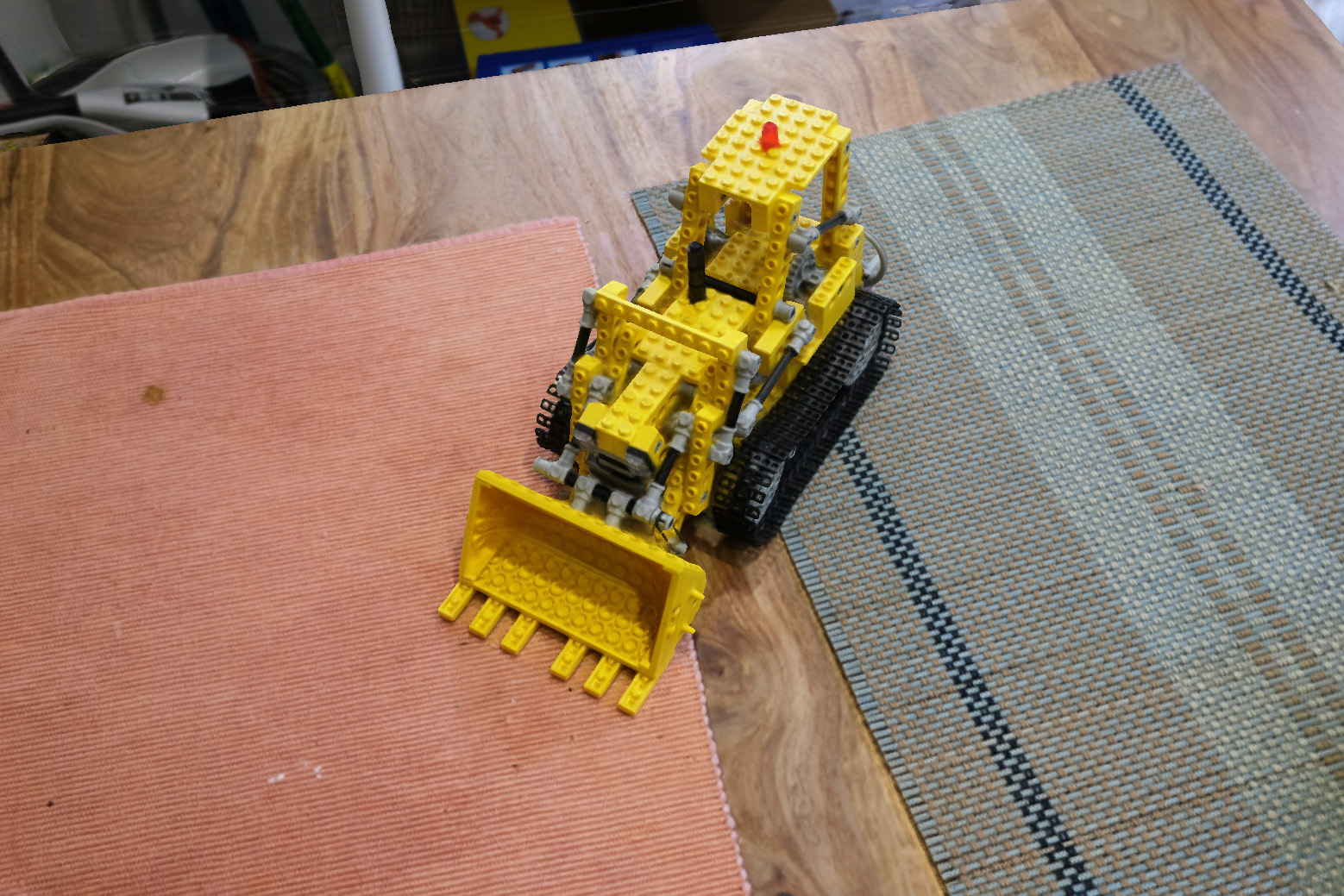} \\
     Mesh & Diffuse color & Specular & Full appearance\\
    \end{tabular}
    \vspace{-10pt}
    \caption{Our method produces an accurate mesh and decomposes appearance into diffuse and specular color.
    }
    \label{fig:appearance}
\end{figure*}

\subsection{Modeling density with an SDF}\label{s:surfacerecon}
Our representation combines the benefits of mip-NeRF 360 for representing unbounded scenes with the well-behaved surface properties of VolSDF's hybrid volume-surface representation~\cite{yariv2021volume}. VolSDF models volumetric density of the scene as a parametric function of an MLP-parameterized signed distance function (SDF) $f$ that returns the signed distance $f(\mathbf{x})$ from each point $\mathbf{x}\in\mathbb{R}^3$ to the surface. 
Because our focus is reconstructing unbounded real-world scenes, we parameterize $f$ in \emph{contracted} space (Equation~\ref{eq:contract}) rather than world-space.
The underlying surface of the scene is the zero-level set of $f$, \ie, the set of points at distance zero from the surface: 
\begin{equation}
    \gS  =  \left\{ \rvx : f(\rvx) = 0 \right\} \, .
\end{equation}
Following VolSDF, we define the volume density $\density$ as:
\begin{equation}\label{eq:density}
\density(\rvx) = \alpha \Psi_\beta \parr{f(\rvx)} \,,
\end{equation}
where $\Psi_\beta$ is the cumulative distribution function of a zero-mean Laplace distribution with scale parameter $\beta > 0$. 
Note that as $\beta$ approaches $0$, the volumetric density approaches a function that returns $\alpha$ inside any object and $0$ in free space.
To encourage $f$ to approximate a valid signed distance function (i.e.\ one where $f(\rvx)$ returns the signed Euclidean distance to the level set of $f$ for all $\rvx$), we penalize the deviation of $f$ from satisfying the Eikonal equation \cite{icml2020_2086}:
\begin{equation}\label{eq:eikonal}
    \gL_{\text{SDF}} = \E_\rvx \left[\left(\norm{\nabla f(\rvx)}-1\right)^2\right] \,.
\end{equation} 

Note that as $f$ is defined in contracted space, this constraint also operates on contracted space.

Recently, Ref-NeRF~\cite{verbin2022ref} improved view-dependent appearance by parameterizing it as a function of the view direction reflected about the surface normal. Our use of an SDF-parameterized density allows this to be easily adopted as SDFs have well-defined surface normals: $\mathbf{n}(\mathbf{x}) = \nabla f(\mathbf{x}) / \norm{\nabla f(\mathbf{x})}$. Therefore, when training this stage of our model we adopt Ref-NeRF's appearance model and compute color using separate diffuse and specular components, where the specular component is parameterized by the concatenation of the view direction reflected about the normal direction, the dot product between the normal and view direction, and a $256$ element bottleneck 
vector output by the MLP that parametrizes $f$. 


We use a variant of mip-NeRF 360 as our model (see Appendix~A in supplementary material for specific training details).
Similarly to VolSDF~\cite{yariv2021volume}, we parameterize the density scale factor as $\alpha = \beta^{-1}$ in Equation~\ref{eq:density}. However, we find that scheduling $\beta$ rather than leaving it as a free optimizable parameter results in more stable training. We therefore anneal $\beta$ according to $\beta_t = \beta_0\left(1 + \frac{\beta_0 - \beta_1}{\beta_1}t^{0.8}\right)^{-1}$,
where $t$ goes from $0$ to $1$ during training, $\beta_0=0.1$, and $\beta_1$ for the three hierarchical sampling stages is $0.015$, $0.003$, and $0.001$ respectively. Because the Eikonal regularization needed for an SDF parameterization of density already removes floaters and results in well-behaved normals, we do not find it necessary to use the orientation loss or predicted normals from Ref-NeRF, or the distortion loss from mip-NeRF 360.





\subsection{Baking a high-resolution mesh}\label{s:bakemesh}
After optimizing our neural volumetric representation, we create a triangle mesh from the recovered MLP-parameterized SDF by querying it on a regular 3D grid and then running Marching Cubes~\cite{lorensen1987marching}.
Note that VolSDF models boundaries using a density fall-off that extends beyond the SDF zero crossing (parameterized by $\beta$). 
We account for this spread when extracting the mesh and choose $0.001$ as the iso-value for surface crossings, as otherwise we find the scene geometry to be slightly eroded.

\paragraph{Visibility and free-space culling} \vspace{-5pt}
When running Marching Cubes,
the MLP-parameterized SDF may contain spurious surface crossings in regions that are occluded from the observed viewpoints as well as regions that the proposal MLP marks as ``free space''. 
The SDF MLP's values in both of these types of regions are not supervised during training, so we must cull any surface crossings that would show up as spurious content in the reconstructed mesh. 
To address this, we inspect the 3D samples taken along the rays in our training data. We compute the volumetric rendering weight for each sample, i.e., how much it contributes to the training pixel color. We then splat any sample with a sufficiently large rendering weight ($>0.005$) into the 3D grid and mark the corresponding cell as a candidate for surface extraction.

\newcommand{\insetwidth}{0.19\linewidth}

\begin{figure*}[ht!]
\centering
\begin{tabular}[!t]{@{}c@{}c@{}}

\makecell[c]{
  \includegraphics[width=0.2\linewidth]{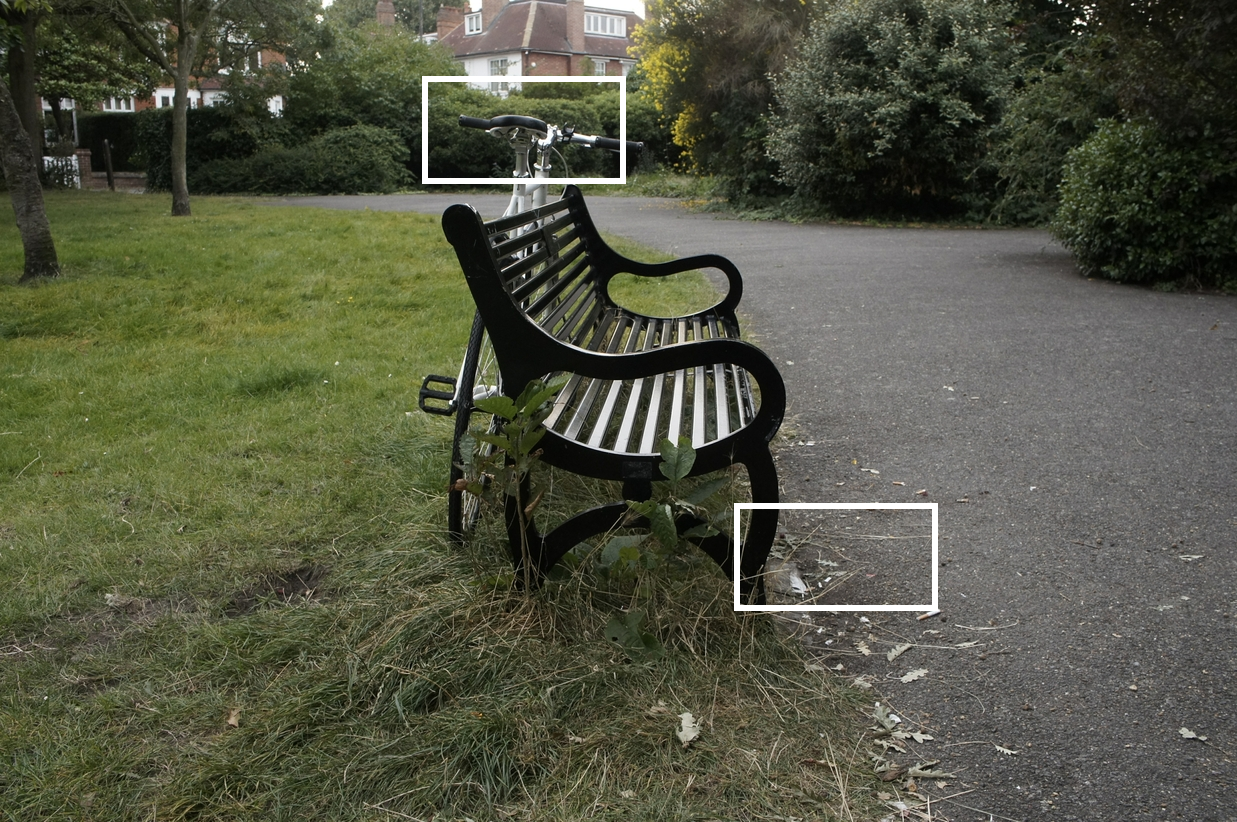}\\
  \texttt{bicycle}
  } & 
  \begin{tabular}[t]{@{\,}c@{\,}c@{\,}c@{\,}c}
  \includegraphics[width=\insetwidth]{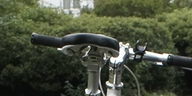} & 
  \includegraphics[width=\insetwidth]{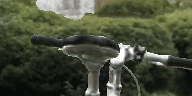} & 
  \includegraphics[width=\insetwidth]{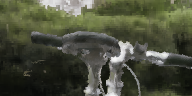} & 
  \includegraphics[width=\insetwidth]{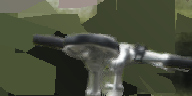} 
  \\
  \includegraphics[width=\insetwidth]{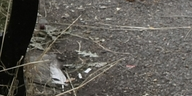} & 
  \includegraphics[width=\insetwidth]{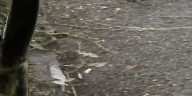} & 
  \includegraphics[width=\insetwidth]{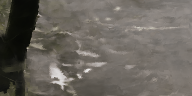} & 
  \includegraphics[width=\insetwidth]{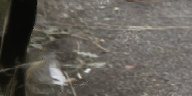} 
  \end{tabular}
 \\
 
\makecell[c]{
  \includegraphics[width=0.2\linewidth]{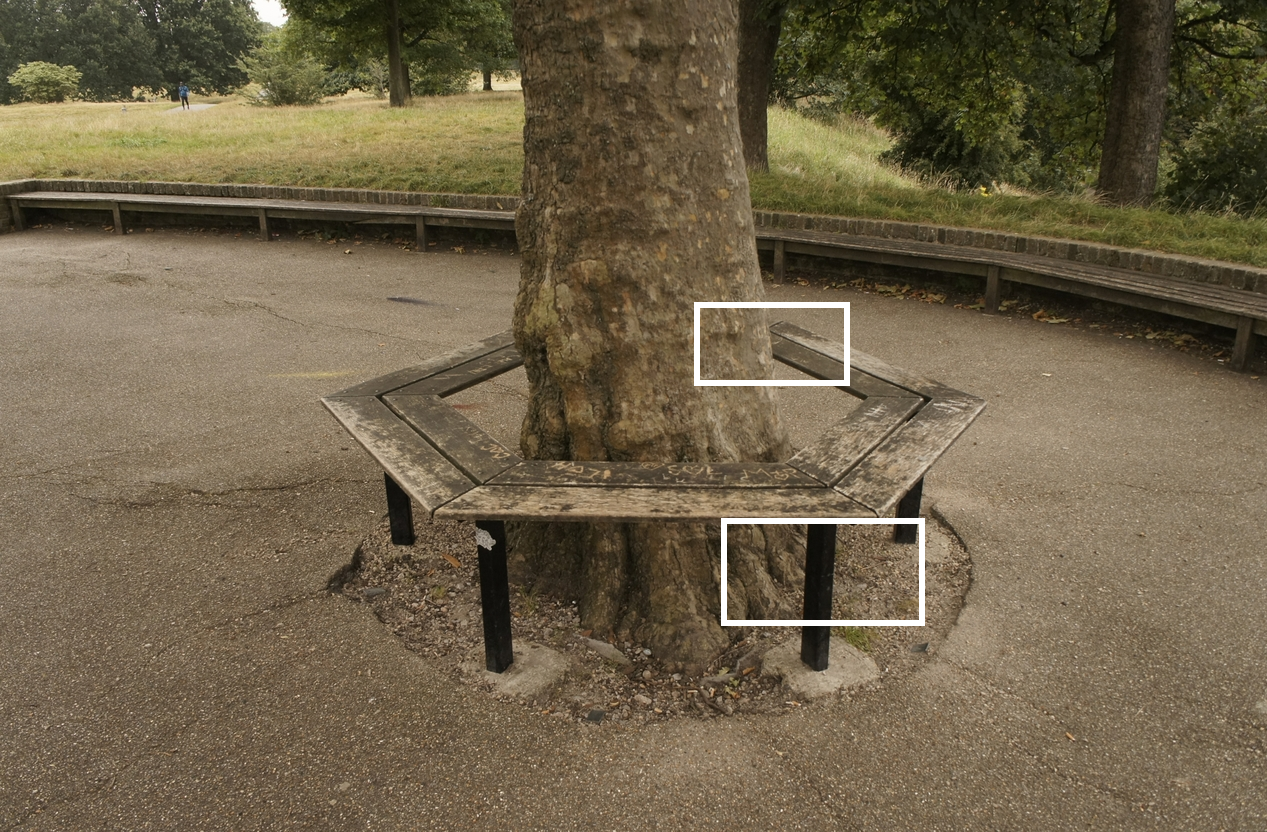}\\
  \texttt{treehill}
  } & 
  \begin{tabular}[t]{@{\,}c@{\,}c@{\,}c@{\,}c}
  \includegraphics[width=\insetwidth]{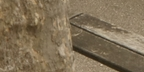} & 
  \includegraphics[width=\insetwidth]{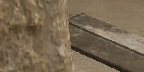} & 
  \includegraphics[width=\insetwidth]{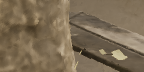} & 
  \includegraphics[width=\insetwidth]{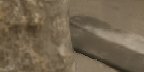} 
  \\
  \includegraphics[width=\insetwidth]{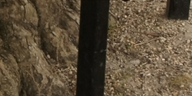} & 
  \includegraphics[width=\insetwidth]{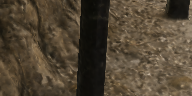} & 
  \includegraphics[width=\insetwidth]{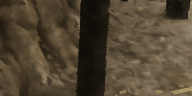} & 
  \includegraphics[width=\insetwidth]{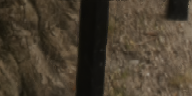} 
  \end{tabular}
 \\
 
\makecell[c]{
  \includegraphics[width=0.2\linewidth]{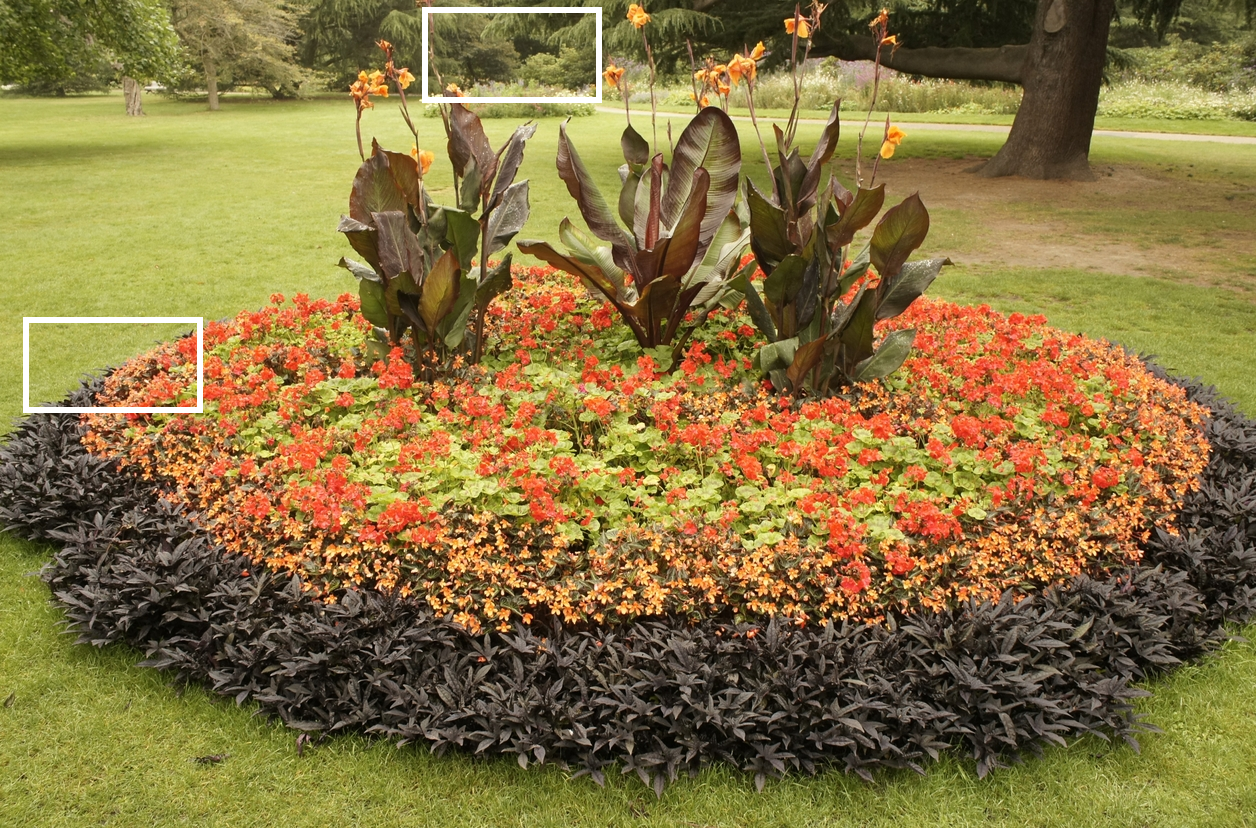}\\
  \texttt{flowerbed}
  } & 
  \begin{tabular}[t]{@{\,}c@{\,}c@{\,}c@{\,}c}
  \includegraphics[width=\insetwidth]{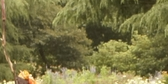} & 
  \includegraphics[width=\insetwidth]{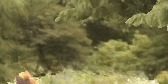} & 
  \includegraphics[width=\insetwidth]{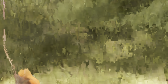} & 
  \includegraphics[width=\insetwidth]{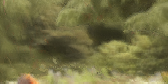} 
  \\
  \includegraphics[width=\insetwidth]{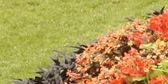} & 
  \includegraphics[width=\insetwidth]{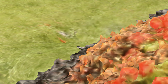} & 
  \includegraphics[width=\insetwidth]{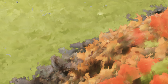} & 
  \includegraphics[width=\insetwidth]{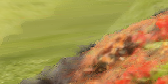} \\
  Ground truth & Ours & Mobile-NeRF & Deep Blending
  \end{tabular}

\end{tabular}
    \vspace{-0.1in}
    \caption{Test-set renderings (with insets) for our model and the two state-of-the-art real-time baselines we evaluate against, using scenes from the mip-NeRF 360 dataset. Deep Blending~\cite{hedman2018deep} produces posterized renderings when the proxy geometry used as input is incorrect (such as in the background of the \texttt{bicycle} scene) and renderings from MobileNeRF~\cite{chen2022mobilenerf} tend to exhibit aliasing artifacts or oversmoothing.} \label{fig:renderings}
\end{figure*}

\paragraph{Mesh extraction} \vspace{-5pt}
We sample our SDF grid at evenly spaced coordinates in the contracted space, which yields unevenly spaced non-axis-aligned coordinates in world space.
This has the desirable property of creating smaller triangles (in world space) for foreground content close to the origin and larger triangles for distant content. Effectively, we leverage the contraction operator as a level-of-detail strategy: as our desired rendered views are close to the scene origin, and because the shape of the contraction is designed to undo the effects of perspective projection, all triangles will have approximately equal areas when projected onto the image plane.

\paragraph{Region growing} \vspace{-5pt}
After extracting the triangle mesh, we use a region growing procedure to fill small holes that might exist in regions that were either unobserved by input viewpoints or missed by the proposal MLP during the baking procedure. We iteratively mark voxels in a neighborhood around the current mesh and extract any surface crossings that exist in these newly active voxels. This \emph{region-growing} strategy effectively remedies situations where a surface exists in the SDF MLP but was not extracted by marching cubes due to insufficient training view coverage or errors in the proposal MLP. We then transform the mesh into world space so it is ready for rasterization by a conventional rendering engine that operates in Euclidean space.

\paragraph{Implementation} \vspace{-5pt}
We use a $2048^3$ grid for both visibility and free-space culling and marching cubes. Initially, we run marching cubes only on voxels that were not culled, i.e. visible and non-empty. We then complete the mesh with 32 region-growing iterations, where we re-run marching cubes in a $8^3$ voxel neighborhood around the vertices in the current mesh. Finally, we post-process the mesh using vertex order optimization~\cite{Sander2007}, which speeds up rendering performance on modern hardware by allowing vertex shader outputs to be cached and reused between neighboring triangles. In Appendix~B we detail additional steps for mesh extraction which do not strictly improve reconstruction accuracy, but enable a more pleasing interactive viewing experience.




\subsection{Modeling view-dependent appearance} \label{s:viewdep} 
The baking procedure described above extracts high-quality triangle mesh geometry from our MLP-based scene representation. To model the scene's appearance, including view-dependent effects such as specularities, we equip each mesh vertex with a diffuse color $\col_d$ and a set of spherical Gaussian lobes. As far-away regions are only observed from a limited set of view directions, we do not need to model view dependence with the same fidelity everywhere in the scene. In our experiments, we use three spherical Gaussian lobes in the central regions ($\norm{\mathbf{x}} \leq 1$) and one lobe in the periphery.
Figure~\ref{fig:appearance} demonstrates our appearance decomposition.

This appearance representation satisfies our efficiency goal for both compute and memory and can thus be rendered in real-time. Each spherical Gaussian lobe has seven parameters: a 3D unit vector $\mu$ for the lobe mean, a 3D vector $\col$ for the lobe color, and a scalar $\lambda$ for the width of the lobe. These lobes are parameterized by the view direction vector $\raydir$, so the rendered color $\mathbf{C}$ for a ray intersecting any given vertex can be computed as:
\begin{equation}\label{eq:sg}
\mathbf{C} = \col_d + \sum_{i=1}^N \col_i \exp\left(\lambda_i \left(\mu_i \cdot \raydir - 1\right) \right) \, .
\end{equation}


\begin{table*}[!]
\centering
\begin{tabular}{l@{\,\,}l|ccc |ccc}
&& \multicolumn{3}{c|}{Outdoor Scenes} & \multicolumn{3}{c}{Indoor Scenes} \\
& & PSNR $\uparrow$ & SSIM $\uparrow$ & LPIPS $\downarrow$ & PSNR $\uparrow$ & SSIM $\uparrow$ & LPIPS $\downarrow$ \\ \hline
\multirow{6}{*}{\rotatebox{90}{offline}} & NeRF \cite{mildenhall2020nerf}                  &                   21.46 &                   0.458 &                   0.515               &                   26.84 &                   0.790 &                   0.370 \\
& NeRF++ \cite{kaizhang2020}             &                   22.76 &                   0.548 &                   0.427  &                   28.05 &                   0.836 &                   0.309 \\
& Stable View Synthesis~\cite{riegler2021stable} & \cellcolor{lightyellow}23.01 & \cellcolor{orange}0.662 &    \cellcolor{tablered}0.253 & 28.22 & \cellcolor{orange}0.907 &    \cellcolor{tablered}0.160\\
& Mip-NeRF 360 \cite{barron2022mipnerf360}          &    \cellcolor{tablered}24.47 &    \cellcolor{tablered}0.691 & \cellcolor{orange}0.283 &    \cellcolor{tablered}31.72 &    \cellcolor{tablered}0.917 & \cellcolor{orange}0.180 \\
& Instant-NGP \cite{muller2022instant}          &                   22.90 &                   0.566 & \cellcolor{lightyellow}0.371 & \cellcolor{lightyellow}29.15 & 0.880 & \cellcolor{lightyellow}0.216 \\
& Ours (offline)        & \cellcolor{orange}23.40 & \cellcolor{lightyellow}0.619 &                   0.379 & \cellcolor{orange}30.21 & \cellcolor{lightyellow}0.888 & 0.243 \\
\hline \hline
\multirow{3}{*}{\rotatebox{90}{real-time}} 
& Deep Blending \cite{hedman2018deep} & \cellcolor{lightyellow} 21.54 & \cellcolor{orange}0.524 & \cellcolor{orange}0.364  & \cellcolor{orange}26.40 & \cellcolor{tablered}0.844 & \cellcolor{orange}0.261 \\
& Mobile-NeRF \cite{chen2022mobilenerf} & \cellcolor{orange}21.95 & \cellcolor{lightyellow}0.470 & \cellcolor{lightyellow}0.470 & $-$ & $-$ & $-$ \\
& Ours (real-time) & \cellcolor{tablered}22.47 &    \cellcolor{tablered}0.585 &    \cellcolor{tablered}0.349 & \cellcolor{tablered}27.06 & \cellcolor{orange}0.836 &    \cellcolor{tablered}0.258 \\
\end{tabular}
\vspace{10pt}
\caption{
Quantitative results of our model on the ``outdoor'' and ``indoor'' scenes from mip-NeRF 360~\cite{barron2022mipnerf360}, with evaluation split for ``offline'' and ``real-time'' algorithms.
Red, orange, and yellow indicate the first, second, and third best performing algorithms for each metric. Metrics not provided by a baseline are denoted with ``$-$''.}
\label{tab:alldoor}
\vspace{-10pt}
\end{table*}



To optimize this representation, we first rasterize the mesh into all training views and store the vertex indices and barycentric coordinates associated with each pixel. 
After this preprocessing, we can easily render a pixel by applying barycentric interpolation to the learned per-vertex parameters and then running our view-dependent appearance model (simulating the operation of a fragment shader).
We can therefore optimize the per-vertex parameters by minimizing a per-pixel color loss as in Equation~\ref{eqn:loss}. As detailed in Appendix~B, we also optimize for a background clear color to provide a more pleasing experience with the interactive viewer. 
To prevent that optimization from being biased by pixels that are not well-modeled by mesh geometry (e.g. pixels at soft object boundaries and semi-transparent objects), instead of the L2 loss that was minimized by VolSDF we use a robust loss $\rho(\cdot, \alpha, c)$ with hyperparameters $\alpha = 0$, $c=\sfrac{1}{5}$ during training, which allows optimization to be more robust to outliers~\cite{BarronCVPR2019}. We also model quantization with a straight-through estimator~\cite{bengio2013estimating}, ensuring that the optimized values for view-dependent appearance are well represented by 8 bits of precision.

We find that directly optimizing this per-vertex representation saturates GPU memory, which prevents us from scaling up to high-resolution meshes. We instead optimize a compressed neural hash-grid model based on Instant NGP~\cite{muller2022instant} (see Appendix A in supplemental material). During optimization, we query this model at each 3D vertex location within a training batch to produce our diffuse colors and spherical Gaussian parameters.

%
%
After optimization is complete, we bake out the compressed scene representation contained in the hash grids by querying the NGP model at each vertex location for the appearance-related parameters. Finally, we export the resulting mesh and per-vertex appearance parameters using the gLTF format~\cite{gltf} and compress it with gzip, a format natively supported by web protocols. 







\section{Experiments}

We evaluate our method's performance both in terms of the accuracy of its output renderings and in terms of its speed, energy, and memory requirements. For accuracy, we test two versions of our model: the intermediate volume rendering results described in Section~\ref{s:surfacerecon}, which we refer to as our ``offline'' model, and the baked real-time model described in Sections~\ref{s:bakemesh} and \ref{s:viewdep}, which we call the ``real-time'' model. As baselines we use prior offline models~\cite{mildenhall2020nerf,kaizhang2020,riegler2021stable,barron2022mipnerf360,muller2022instant} designed for fidelity, as well as with prior real-time methods~\cite{hedman2018deep,chen2022mobilenerf} designed for performance. We additionally compare our method's recovered meshes with those extracted by COLMAP~\cite{schoenberger2016mvs}, mip-NeRF 360~\cite{barron2022mipnerf360}, and MobileNeRF~\cite{chen2022mobilenerf}. 
All FPS (frames-per-second) measurements are for rendering at $1920\times 1080$ resolution.

\vspace{-5pt}
\subsection{Real-time rendering of unbounded scenes}

We evaluate our method on the dataset of real-world scenes from mip-NeRF 360~\cite{barron2022mipnerf360}, which contains complicated indoor and outdoor scenes captured from all viewing angles.
In Table~\ref{tab:alldoor} we present a quantitative evaluation of both the offline and real-time versions of our model against our baselines. Though our offline model is outperformed by some prior works (as we might expect, given that our focus is performance) our real-time method outperforms the two recent state-of-the-art real-time baselines we evaluate again across all three error metrics used by this benchmark.
In Figure~\ref{fig:renderings} we show a qualitative comparison of renderings from our model and these two state-of-the-art real-time baselines, and we observe that our approach exhibits more detail and fewer artifacts than prior work.

In Table~\ref{tab:perf} we evaluate our method's rendering performance by comparing against Instant-NGP (the fastest ``offline'' model we evaluate against) and MobileNeRF (the real-time model that produces the highest quality renderings after our own). We measure performance of all methods at $1920\times 1080$. Both MobileNeRF and our method are running in-browser on a 16" Macbook Pro with a Radeon 5500M GPU while Instant NGP is running on a workstation equipped with a power NVIDIA RTX 3090 GPU.
Though our approach requires more on-disk storage than MobileNeRF ($1.27 \times$) and Instant NGP ($4.07 \times$), we see that our model is significantly more efficient than both baselines --- our model yields FPS/Watt metrics that are $1.44 \times$ and $77 \times$ greater respectively, in addition to producing higher quality renderings. 


\begin{table}[h]
\resizebox{\linewidth}{!}{
\begin{tabular}{@{}l|cccc@{}}
& W $\downarrow$ & FPS $\uparrow$ & FPS/W $\uparrow$  & MB (disk) $\downarrow$ \\ \hline
Instant-NGP \cite{muller2022instant} & 350 & 3.78 & 0.011 &    \textbf{106.8} \\
Mobile-NeRF \cite{chen2022mobilenerf} & \textbf{85} & 50.06 & 0.589 & 341.9 \\
Ours        &    \textbf{85} &    \textbf{72.21} &    \textbf{0.850} & 434.5 \\
\end{tabular}
}
\vspace{5pt}
\caption{
The performance (Watts consumed, frames per second, and their ratio) and storage requirements for our real-time method and two baselines. FPS is measured when rendering at $1920\times 1080$ resolution.} 
\label{tab:perf}
\end{table}

Our appreciably improved performance relative to MobileNeRF may seem unusual at first glance, as both our approach and MobileNeRF both yield optimized meshes that can be easily and quickly rasterized. This discrepancy is likely due to MobileNeRF's reliance on alpha masking (which results in a significant amount of compute-intensive overdraw) and MobileNeRF's use of an MLP to model view-dependent radiance (which requires more compute to evaluate than our spherical Gaussian approach).

Compared to Deep Blending~\cite{hedman2018deep}, we see from Table~\ref{tab:alldoor} that our method achieves higher quality. However, it is also worth noting that our representation is also much \emph{simpler}: while our meshes can be rendered in a browser, Deep Blending relies on carefully tuned CUDA rendering and must store both color and geometry for all training images in the scene. As a result, total storage cost for Deep Blending in the outdoor scenes is $2.66\times$ higher (1154.78 MB on average) than for our corresponding meshes.




\begin{figure}[!ht]
\vspace{10pt}
    \centering
    \begin{tabular}[width=\linewidth]{@{}c@{\,\,}c@{\,\,}c@{\,\,}c@{}}
    COLMAP & MobileNeRF &  Mip-NeRF360 & Ours \\
     \includegraphics[width=0.24\linewidth]{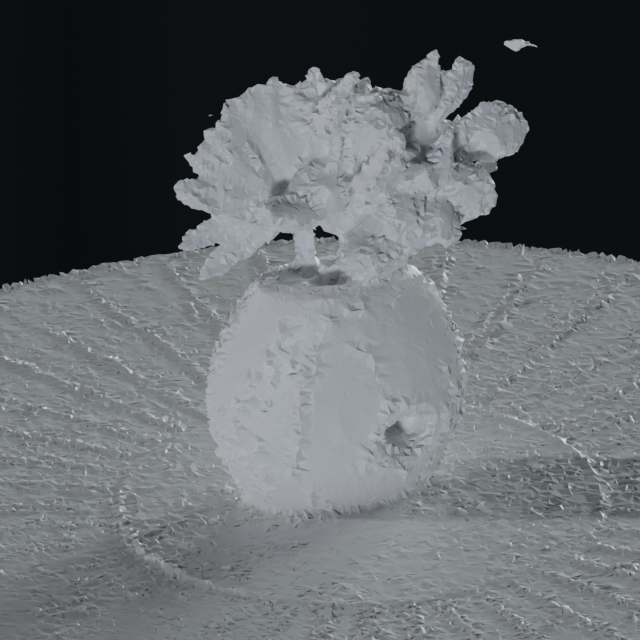} & 
     \includegraphics[width=0.24\linewidth]{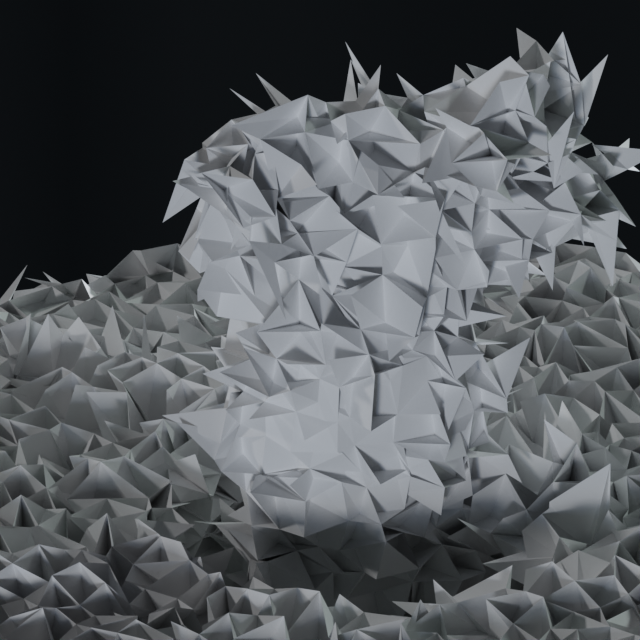} & 
     \includegraphics[width=0.24\linewidth]{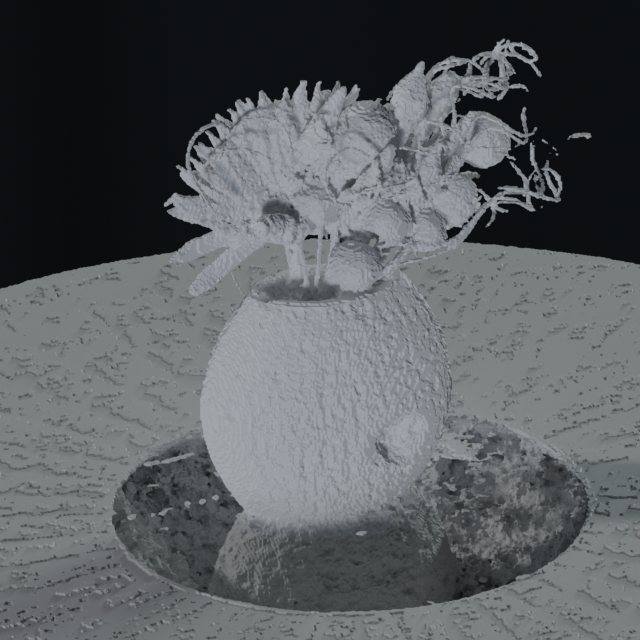} & 
     \includegraphics[width=0.24\linewidth]{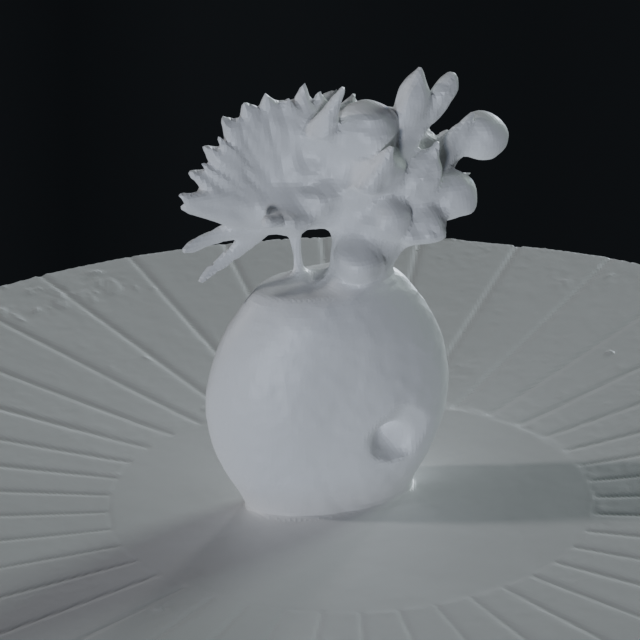} \\
     \includegraphics[width=0.24\linewidth]{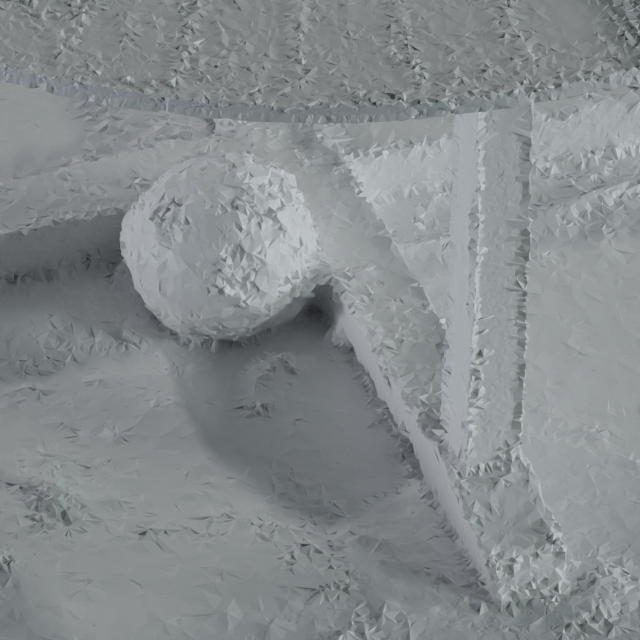} & 
     \includegraphics[width=0.24\linewidth]{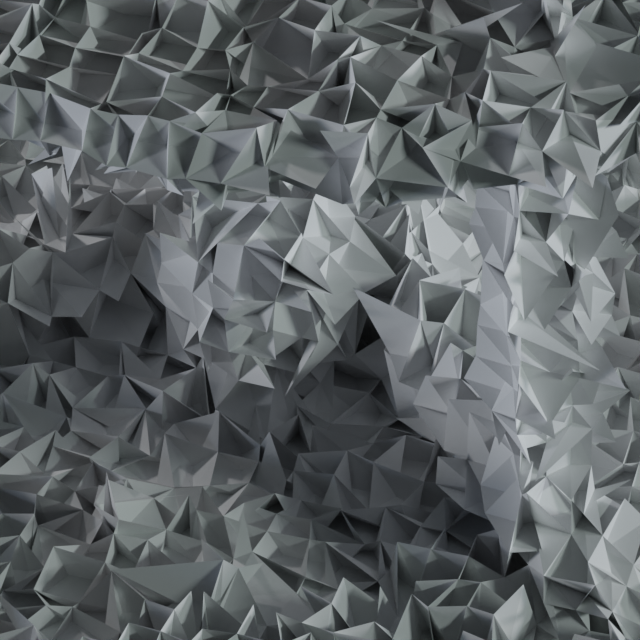} & 
     \includegraphics[width=0.24\linewidth]{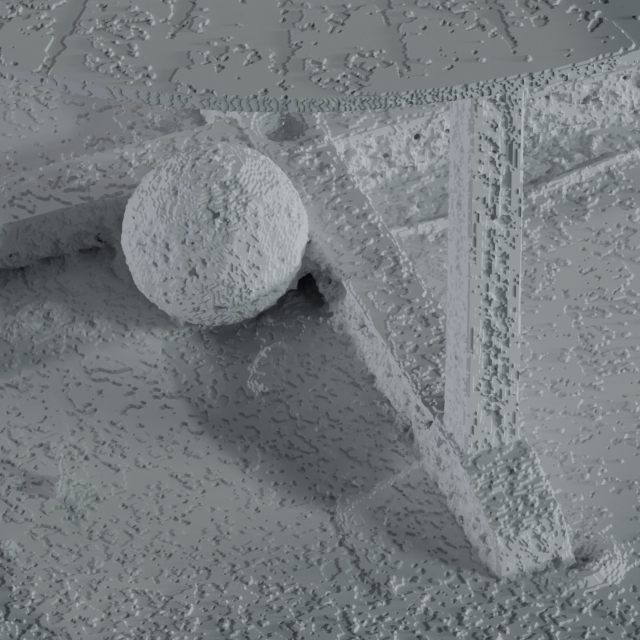} & 
     \includegraphics[width=0.24\linewidth]{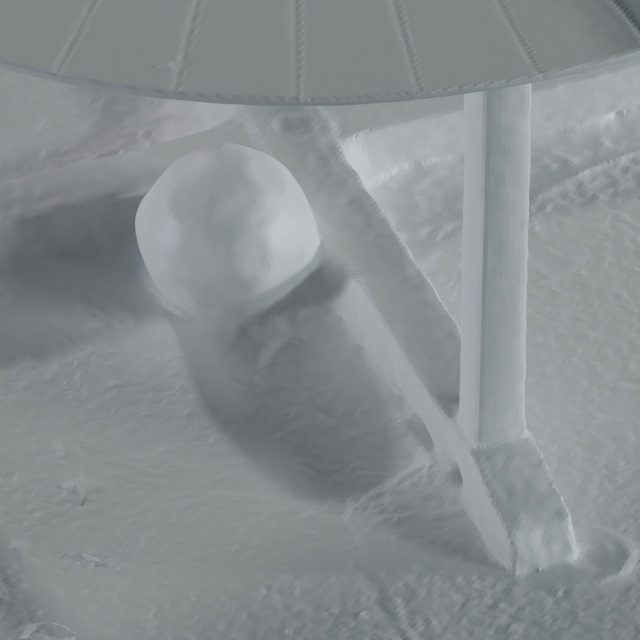}
    \end{tabular}
    \vspace{-10pt}
    \caption{Comparing the meshes produced by our technique with baselines that yield meshes. 
    Our meshes are higher in quality compared to those of COLMAP, MobileNeRF, and Mip-NeRF 360. COLMAP's mesh contains noise, floaters, and irregular object boundaries, MobileNeRF's mesh is a ``polygon soup'' that may not accurately represent scene geometry, and iso-surfaces from Mip-NeRF 360's density field tend to be noisy and represent reflections with inaccurate geometry.
    }
    \label{fig:meshes}
\end{figure}

\subsection{Mesh extraction}

In Figure~\ref{fig:meshes} we present a qualitative comparison of our mesh with those obtained using  COLMAP~\cite{schoenberger2016mvs}, MobileNeRF~\cite{chen2022mobilenerf} and an iso-surface of Mip-NeRF 360 \cite{barron2022mipnerf360}. We evaluate against COLMAP not only because it represents a mature structure-from-motion software package, but also because the geometry produced by COLMAP is used as input by Stable View Synthesis and Deep Blending.
COLMAP uses volumetric graph cuts on a tetrahedralization of the scene~\cite{labatut2007efficient,jancosek2011multi} to obtain a binary segmentation of the scene and then forms a triangle mesh as the surface between these regions. Because this binary segmentation does not allow for any averaging of the surface, small noise in the initial reconstruction tends to result in noisy reconstructed meshes, which results in a ``bumpy'' appearance.
MobileNeRF represents the scene as a disconnected collection of triangles, as its sole focus is view synthesis. As a result, its optimized and pruned ``triangle soup'' is highly noisy and may not be ideal for downstream tasks such as appearance editing. 

As recently shown \cite{oechsle2021unisurf, yariv2021volume, wang2021neus}, extracting an iso-surface directly from the density field predicted by NeRF can sometimes fail to faithfully capture the geometry of the scene. 
In Figure~\ref{fig:meshes} we show this effect using Mip-NeRF 360 and extract the iso-surface where its density field exceeds a value of 50. Note how the surface of the table is no longer flat, as the reflection of the vase is modeled using mirror-world geometry.
In contrast, our method produces a smooth and high-fidelity mesh, which is better suited for appearance and illumination editing, as demonstrated in Figure~\ref{fig:teaser}.

\vspace{15pt}
\subsection{Appearance model ablation}

 In Table~\ref{tab:ablationappearance} we present the results of an ablation study of our spherical Gaussian appearance model. We see that reducing the number of SGs to 2, 1, and 0 (\ie, a diffuse model) causes accuracy to degrade monotonically. However, when using 3 SGs in the periphery our model tends to overfit to the training views, causing a slight drop in quality compared to our proposed model with just a single peripheral SG. Furthermore, compared to 3 SGs everywhere, using a single SG in the periphery reduces the average size vertex by $1.52\times$ (from $36$ to $23.76$ bytes), which significantly reduces the memory bandwidth consumption (a major performance bottleneck for rendering).
 Perhaps surprisingly, replacing our SG appearance model with the small view-dependent MLP used by both SNeRG~\cite{hedman2021snerg} and MobileNeRF~\cite{chen2022mobilenerf} significantly reduces rendering quality and yields error metrics that are roughly comparable to the ``1 Spherical Gaussian'' ablation. This is especially counter-intuitive given the significant cost of evaluating a small MLP ($\sim\!2070$ FLOPS per pixel) compared to a single spherical Gaussian ($21$ FLOPS per pixel). Additionally, we ablate the robust loss used to train our appearance representation with a simple L2 loss, which unsurprisingly boosts PSNR (which is inversely proportional to MSE) at the expense of the other metrics.
\vspace{3pt} 

 \begin{table}[h]
\resizebox{\linewidth}{!}{
\begin{tabular}{@{}l|cccc}
& PSNR $\uparrow$ & SSIM $\uparrow$ & LPIPS $\downarrow$ & MB (GPU) $\downarrow$ \\ \hline
Diffuse (0 Spherical Gaussians) &  22.32 & 0.636 &   0.352 & 436.1 \\ \hdashline 
1 Spherical Gaussian   &       24.02 &        0.680 &      0.322 & \cellcolor{orange}549.1 \\
2 Spherical Gaussian     & \cellcolor{lightyellow}24.39 & \cellcolor{orange}0.693 & \cellcolor{orange}0.312  & 662.2 \\
3 SGs in the periphery     &    24.34 &     0.688 &    0.317  & 775.3\\
View-dependent MLP~\shortcite{hedman2021snerg}              &                   24.30 &                   0.687 &                   0.318 & \cellcolor{tablered}516.8 \\
L2 loss   &    \cellcolor{tablered}24.52 & \cellcolor{lightyellow}0.690 & \cellcolor{lightyellow}0.316 & \cellcolor{lightyellow}572.6  \\
Ours     & \cellcolor{orange}24.51 &    \cellcolor{tablered}0.697 &    \cellcolor{tablered}0.309 & \cellcolor{lightyellow}572.6  \\
\end{tabular}
}
\caption{
An ablation study of our view-dependent appearance model on all scenes from the mip-NeRF 360 dataset.}
\label{tab:ablationappearance}
\end{table}

 



\begin{figure}[!ht]
    \centering
    \begin{tabular}[width=\linewidth]{@{}c@{\,\,}c@{\,\,}c@{}}
    \includegraphics[width=0.32\linewidth]{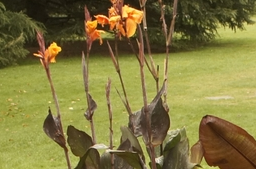} &
    \includegraphics[width=0.32\linewidth]{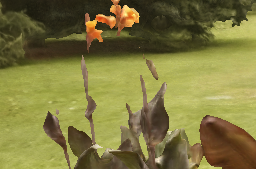} &
    \includegraphics[width=0.32\linewidth]{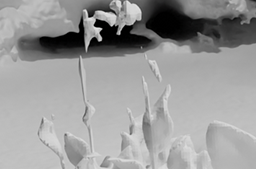} \\
    \includegraphics[width=0.32\linewidth]{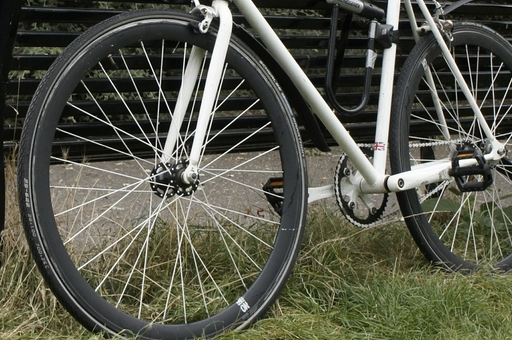} &
    \includegraphics[width=0.32\linewidth]{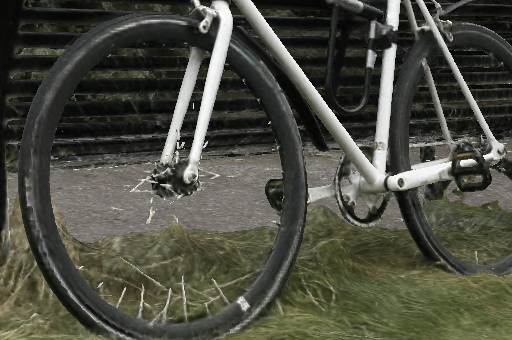} &
    \includegraphics[width=0.32\linewidth]{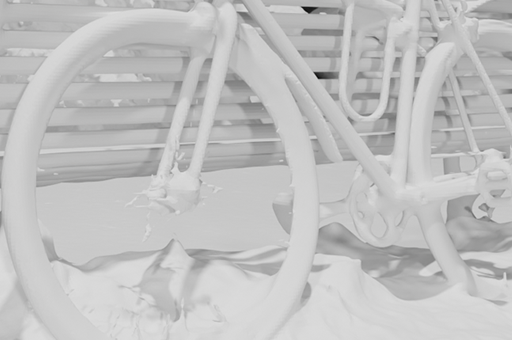} \\
    \includegraphics[width=0.32\linewidth]{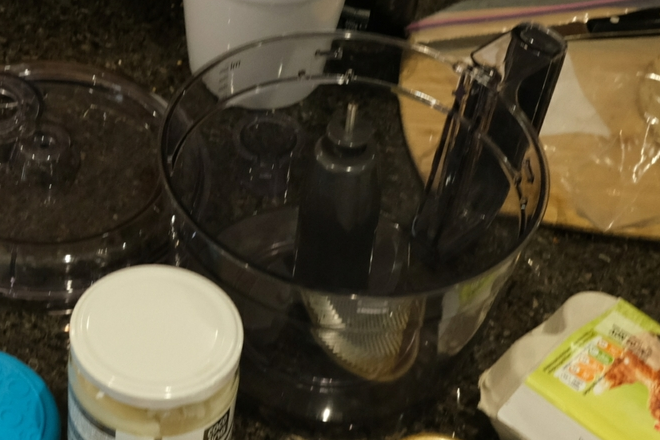} &
    \includegraphics[width=0.32\linewidth]{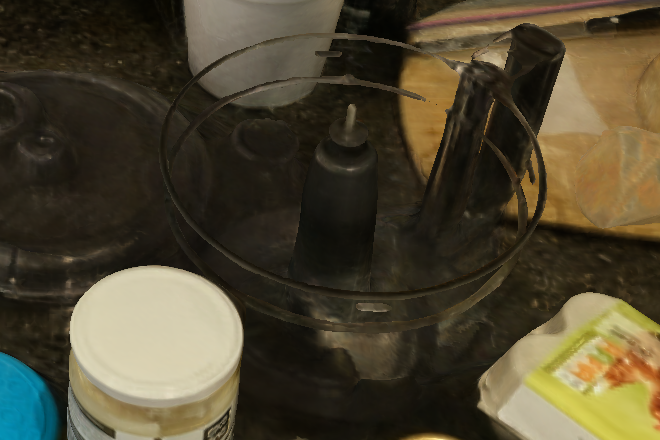} &
    \includegraphics[width=0.32\linewidth]{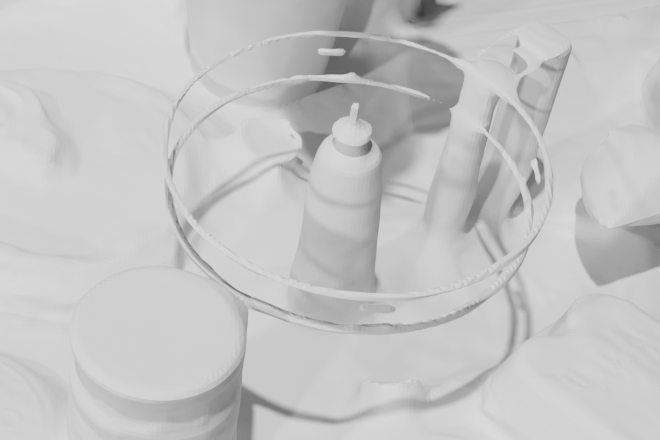} \\
    Ground truth & Our rendering &  Our mesh
    \end{tabular}
    \vspace{-10pt}
    \caption{Our framework is based on the neural SDF representation, which struggles to represent semi-transparent objects or thin structures. These limitations can further affect our rendering reconstruction performance.
    }
    \label{fig:limitation}
\end{figure}

\vspace{-8pt}
\subsection{Limitations}
Although our model achieves state-of-the-art speed and accuracy for the established task of real-time rendering of unbounded scenes, there are several limitations that represent opportunities for future improvement: We represent the scene using a fully opaque mesh representation, and as such our model may struggle to represent semi-transparent content (glass, fog, etc.). And as is common for mesh-based approaches, our model sometimes fails to accurately represent areas with small or detailed geometry (dense foliage, thin structures, etc.). Figure~\ref{fig:limitation} depicts additional extracted meshes visualization which demonstrates our surface reconstruction limitations and their effect on the rendering reconstruction. These concerns could perhaps be addressed by augmenting the mesh with opacity values, but allowing for continuous opacity would require a complex polygon sorting procedure that is difficult to integrate into a real-time rasterization pipeline. One additional limitation of our technique is that our model's output meshes occupy a significant amount of on-disk space ($\sim\!430$ megabytes per scene), which may prove challenging to store or stream for some applications. This could be ameliorated through mesh simplification followed by UV atlasing. However, we found that existing tools for simplification and atlasing, which are mostly designed for artist-made 3D assets, did not work well for our meshes extracted by marching cubes.



\section{Conclusion}

We have presented a system that produces a high-quality mesh for real-time rendering of large unbounded real world scenes. Our technique first optimizes a hybrid neural volume-surface representation of the scene that is designed for accurate surface reconstruction. From this hybrid representation, we extract a triangle mesh whose vertices contain an efficient representation of view-dependent appearance, then optimize this meshed representation to best reproduce the captured input images. This results in a mesh that yields state-of-the-art results for real-time view synthesis in terms of both speed and in accuracy, and is of a high enough quality to enable downstream applications.

\begin{acks}
We would like to thank Forrester Cole and Srinivas Kaza for their implementation of the JAX rasterizer, Simon Rodriguez as an invaluable source of knowledge for real-time graphics programming, and Marcos Seefelder for brainstorming the real-time renderer. We further thank Thomas Müller for his valuable advice on tuning Instant-NGP for the Mip-NeRF 360 dataset, and Zhiqin Chen for generously sharing with us the MobileNeRF evaluations. Lastly, we thank Keunhong Park for thoughtful review of our manuscript.




\end{acks}


\bibliographystyle{ACM-Reference-Format}
\bibliography{sample-bibliography}


\begin{thebibliography}{58}


\ifx \showCODEN    \undefined \def \showCODEN     #1{\unskip}     \fi
\ifx \showDOI      \undefined \def \showDOI       #1{#1}\fi
\ifx \showISBNx    \undefined \def \showISBNx     #1{\unskip}     \fi
\ifx \showISBNxiii \undefined \def \showISBNxiii  #1{\unskip}     \fi
\ifx \showISSN     \undefined \def \showISSN      #1{\unskip}     \fi
\ifx \showLCCN     \undefined \def \showLCCN      #1{\unskip}     \fi
\ifx \shownote     \undefined \def \shownote      #1{#1}          \fi
\ifx \showarticletitle \undefined \def \showarticletitle #1{#1}   \fi
\ifx \showURL      \undefined \def \showURL       {\relax}        \fi
\providecommand\bibfield[2]{#2}
\providecommand\bibinfo[2]{#2}
\providecommand\natexlab[1]{#1}
\providecommand\showeprint[2][]{arXiv:#2}

\bibitem[Barron(2019)]%
        {BarronCVPR2019}
\bibfield{author}{\bibinfo{person}{Jonathan~T. Barron}.}
  \bibinfo{year}{2019}\natexlab{}.
\newblock \showarticletitle{A General and Adaptive Robust Loss Function}.
\newblock \bibinfo{journal}{\emph{CVPR}} (\bibinfo{year}{2019}).
\newblock


\bibitem[Barron et~al\mbox{.}(2022)]%
        {barron2022mipnerf360}
\bibfield{author}{\bibinfo{person}{Jonathan~T. Barron}, \bibinfo{person}{Ben
  Mildenhall}, \bibinfo{person}{Dor Verbin}, \bibinfo{person}{Pratul~P.
  Srinivasan}, {and} \bibinfo{person}{Peter Hedman}.}
  \bibinfo{year}{2022}\natexlab{}.
\newblock \showarticletitle{{Mip-NeRF 360}: Unbounded Anti-Aliased Neural
  Radiance Fields}.
\newblock \bibinfo{journal}{\emph{CVPR}} (\bibinfo{year}{2022}).
\newblock


\bibitem[Bengio et~al\mbox{.}(2013)]%
        {bengio2013estimating}
\bibfield{author}{\bibinfo{person}{Yoshua Bengio}, \bibinfo{person}{Nicholas
  L{\'e}onard}, {and} \bibinfo{person}{Aaron Courville}.}
  \bibinfo{year}{2013}\natexlab{}.
\newblock \showarticletitle{Estimating or propagating gradients through
  stochastic neurons for conditional computation}.
\newblock \bibinfo{journal}{\emph{arXiv preprint arXiv:1308.3432}}
  (\bibinfo{year}{2013}).
\newblock


\bibitem[Boss et~al\mbox{.}(2021)]%
        {boss2021nerd}
\bibfield{author}{\bibinfo{person}{Mark Boss}, \bibinfo{person}{Raphael Braun},
  \bibinfo{person}{Varun Jampani}, \bibinfo{person}{Jonathan~T. Barron},
  \bibinfo{person}{Ce Liu}, {and} \bibinfo{person}{Hendrik P.~A. Lensch}.}
  \bibinfo{year}{2021}\natexlab{}.
\newblock \showarticletitle{{NeRD}: Neural Reflectance Decomposition from Image
  Collections}.
\newblock \bibinfo{journal}{\emph{ICCV}} (\bibinfo{year}{2021}).
\newblock


\bibitem[Boss et~al\mbox{.}(2022)]%
        {boss2022-samurai}
\bibfield{author}{\bibinfo{person}{Mark Boss}, \bibinfo{person}{Andreas
  Engelhardt}, \bibinfo{person}{Abhishek Kar}, \bibinfo{person}{Yuanzhen Li},
  \bibinfo{person}{Deqing Sun}, \bibinfo{person}{Jonathan~T. Barron},
  \bibinfo{person}{Hendrik~P.A. Lensch}, {and} \bibinfo{person}{Varun
  Jampani}.} \bibinfo{year}{2022}\natexlab{}.
\newblock \showarticletitle{{SAMURAI}: {S}hape {A}nd {M}aterial from
  {U}nconstrained {R}eal-world {A}rbitrary {I}mage collections}.
\newblock \bibinfo{journal}{\emph{NeurIPS}} (\bibinfo{year}{2022}).
\newblock


\bibitem[Buehler et~al\mbox{.}(2001)]%
        {buehler2001unstructured}
\bibfield{author}{\bibinfo{person}{Chris Buehler}, \bibinfo{person}{Michael
  Bosse}, \bibinfo{person}{Leonard McMillan}, \bibinfo{person}{Steven Gortler},
  {and} \bibinfo{person}{Michael Cohen}.} \bibinfo{year}{2001}\natexlab{}.
\newblock \showarticletitle{Unstructured Lumigraph Rendering}.
\newblock \bibinfo{journal}{\emph{SIGGRAPH}} (\bibinfo{year}{2001}).
\newblock


\bibitem[Chen et~al\mbox{.}(2022b)]%
        {chen2022tensorf}
\bibfield{author}{\bibinfo{person}{Anpei Chen}, \bibinfo{person}{Zexiang Xu},
  \bibinfo{person}{Andreas Geiger}, \bibinfo{person}{Jingyi Yu}, {and}
  \bibinfo{person}{Hao Su}.} \bibinfo{year}{2022}\natexlab{b}.
\newblock \showarticletitle{{TensoRF}: Tensorial Radiance Fields}.
\newblock \bibinfo{journal}{\emph{ECCV}} (\bibinfo{year}{2022}).
\newblock


\bibitem[Chen et~al\mbox{.}(2022a)]%
        {chen2022mobilenerf}
\bibfield{author}{\bibinfo{person}{Zhiqin Chen}, \bibinfo{person}{Thomas
  Funkhouser}, \bibinfo{person}{Peter Hedman}, {and} \bibinfo{person}{Andrea
  Tagliasacchi}.} \bibinfo{year}{2022}\natexlab{a}.
\newblock \showarticletitle{{MobileNeRF}: Exploiting the polygon rasterization
  pipeline for efficient neural field rendering on mobile architectures}.
\newblock \bibinfo{journal}{\emph{arXiv:2208.00277}} (\bibinfo{year}{2022}).
\newblock


\bibitem[Debevec et~al\mbox{.}(1996)]%
        {debevec:1996:vdtm}
\bibfield{author}{\bibinfo{person}{Paul~E. Debevec},
  \bibinfo{person}{Camillo~J. Taylor}, {and} \bibinfo{person}{Jitendra Malik}.}
  \bibinfo{year}{1996}\natexlab{}.
\newblock \showarticletitle{Modeling and Rendering Architecture from
  Photographs: A hybrid geometry- and image-based approach}.
\newblock \bibinfo{journal}{\emph{SIGGRAPH}} (\bibinfo{year}{1996}).
\newblock


\bibitem[Furukawa and Hern\'andez(2015)]%
        {Furukawa_Fnt15}
\bibfield{author}{\bibinfo{person}{Yasutaka Furukawa} {and}
  \bibinfo{person}{Carlos Hern\'andez}.} \bibinfo{year}{2015}\natexlab{}.
\newblock \showarticletitle{Multi-View Stereo: A Tutorial}.
\newblock \bibinfo{journal}{\emph{Foundations and Trends in Computer Graphics
  and Vision}} (\bibinfo{year}{2015}).
\newblock


\bibitem[Garbin et~al\mbox{.}(2021)]%
        {garbin2021fastnerf}
\bibfield{author}{\bibinfo{person}{Stephan~J. Garbin}, \bibinfo{person}{Marek
  Kowalski}, \bibinfo{person}{Matthew Johnson}, \bibinfo{person}{Jamie
  Shotton}, {and} \bibinfo{person}{Julien Valentin}.}
  \bibinfo{year}{2021}\natexlab{}.
\newblock \showarticletitle{{FastNeRF}: High-Fidelity Neural Rendering at
  200FPS}.
\newblock \bibinfo{journal}{\emph{ICCV}} (\bibinfo{year}{2021}).
\newblock


\bibitem[Gortler et~al\mbox{.}(1996)]%
        {gortler1996lumigraph}
\bibfield{author}{\bibinfo{person}{Steven~J. Gortler}, \bibinfo{person}{Radek
  Grzeszczuk}, \bibinfo{person}{Richard Szeliski}, {and}
  \bibinfo{person}{Michael~F. Cohen}.} \bibinfo{year}{1996}\natexlab{}.
\newblock \showarticletitle{The lumigraph}.
\newblock \bibinfo{journal}{\emph{SIGGRAPH}} (\bibinfo{year}{1996}).
\newblock


\bibitem[Gropp et~al\mbox{.}(2020)]%
        {icml2020_2086}
\bibfield{author}{\bibinfo{person}{Amos Gropp}, \bibinfo{person}{Lior Yariv},
  \bibinfo{person}{Niv Haim}, \bibinfo{person}{Matan Atzmon}, {and}
  \bibinfo{person}{Yaron Lipman}.} \bibinfo{year}{2020}\natexlab{}.
\newblock \showarticletitle{Implicit Geometric Regularization for Learning
  Shapes}.
\newblock \bibinfo{journal}{\emph{Proceedings of Machine Learning and Systems}}
  (\bibinfo{year}{2020}).
\newblock


\bibitem[Hedman et~al\mbox{.}(2018)]%
        {hedman2018deep}
\bibfield{author}{\bibinfo{person}{Peter Hedman}, \bibinfo{person}{Julien
  Philip}, \bibinfo{person}{True Price}, \bibinfo{person}{Jan-Michael Frahm},
  \bibinfo{person}{George Drettakis}, {and} \bibinfo{person}{Gabriel Brostow}.}
  \bibinfo{year}{2018}\natexlab{}.
\newblock \showarticletitle{Deep blending for free-viewpoint image-based
  rendering}.
\newblock \bibinfo{journal}{\emph{SIGGRAPH Asia}} (\bibinfo{year}{2018}).
\newblock


\bibitem[Hedman et~al\mbox{.}(2021)]%
        {hedman2021snerg}
\bibfield{author}{\bibinfo{person}{Peter Hedman}, \bibinfo{person}{Pratul~P.
  Srinivasan}, \bibinfo{person}{Ben Mildenhall}, \bibinfo{person}{Jonathan~T.
  Barron}, {and} \bibinfo{person}{Paul Debevec}.}
  \bibinfo{year}{2021}\natexlab{}.
\newblock \showarticletitle{Baking Neural Radiance Fields for Real-Time View
  Synthesis}.
\newblock \bibinfo{journal}{\emph{ICCV}} (\bibinfo{year}{2021}).
\newblock


\bibitem[Hendrycks and Gimpel(2016)]%
        {hendrycks2016gaussian}
\bibfield{author}{\bibinfo{person}{Dan Hendrycks} {and} \bibinfo{person}{Kevin
  Gimpel}.} \bibinfo{year}{2016}\natexlab{}.
\newblock \showarticletitle{Gaussian Error Linear Units ({GELUs})}.
\newblock \bibinfo{journal}{\emph{arXiv:1606.08415}} (\bibinfo{year}{2016}).
\newblock


\bibitem[ISO/IEC 12113:2022(2022)]%
        {gltf}
ISO/IEC 12113:2022 \bibinfo{year}{2022}\natexlab{}.
\newblock \bibinfo{booktitle}{\emph{{Information technology — Runtime 3D
  asset delivery format — Khronos glTF 2.0}}}.
\newblock \bibinfo{type}{Standard}. \bibinfo{institution}{International
  Organization for Standardization}.
\newblock


\bibitem[Jancosek and Pajdla(2011)]%
        {jancosek2011multi}
\bibfield{author}{\bibinfo{person}{Michal Jancosek} {and}
  \bibinfo{person}{Tom{\'a}s Pajdla}.} \bibinfo{year}{2011}\natexlab{}.
\newblock \showarticletitle{Multi-view reconstruction preserving
  weakly-supported surfaces}.
\newblock  (\bibinfo{year}{2011}).
\newblock


\bibitem[Karnewar et~al\mbox{.}(2022)]%
        {karnewar2022relu}
\bibfield{author}{\bibinfo{person}{Animesh Karnewar}, \bibinfo{person}{Tobias
  Ritschel}, \bibinfo{person}{Oliver Wang}, {and} \bibinfo{person}{Niloy
  Mitra}.} \bibinfo{year}{2022}\natexlab{}.
\newblock \showarticletitle{{ReLU} fields: The little non-linearity that
  could}.
\newblock \bibinfo{journal}{\emph{SIGGRAPH}} (\bibinfo{year}{2022}).
\newblock


\bibitem[Kazhdan et~al\mbox{.}(2006)]%
        {SGP:SGP06:061-070}
\bibfield{author}{\bibinfo{person}{Michael Kazhdan}, \bibinfo{person}{Matthew
  Bolitho}, {and} \bibinfo{person}{Hugues Hoppe}.}
  \bibinfo{year}{2006}\natexlab{}.
\newblock \showarticletitle{Poisson Surface Reconstruction}.
\newblock \bibinfo{journal}{\emph{Symposium on Geometry Processing}}
  (\bibinfo{year}{2006}).
\newblock
\showISBNx{3-905673-24-X}
\showISSN{1727-8384}


\bibitem[Kazhdan and Hoppe(2013)]%
        {kazhdan2013screened}
\bibfield{author}{\bibinfo{person}{Michael Kazhdan} {and}
  \bibinfo{person}{Hugues Hoppe}.} \bibinfo{year}{2013}\natexlab{}.
\newblock \showarticletitle{Screened {Poisson} Surface Reconstruction}.
\newblock \bibinfo{journal}{\emph{ACM TOG}} (\bibinfo{year}{2013}).
\newblock


\bibitem[Kingma and Ba(2015)]%
        {adam}
\bibfield{author}{\bibinfo{person}{Diederik~P. Kingma} {and}
  \bibinfo{person}{Jimmy Ba}.} \bibinfo{year}{2015}\natexlab{}.
\newblock \showarticletitle{Adam: A method for stochastic optimization}.
\newblock \bibinfo{journal}{\emph{ICLR}} (\bibinfo{year}{2015}).
\newblock


\bibitem[Kopanas et~al\mbox{.}(2021)]%
        {kopanas2021point}
\bibfield{author}{\bibinfo{person}{Georgios Kopanas}, \bibinfo{person}{Julien
  Philip}, \bibinfo{person}{Thomas Leimk{\"u}hler}, {and}
  \bibinfo{person}{George Drettakis}.} \bibinfo{year}{2021}\natexlab{}.
\newblock \showarticletitle{Point-Based Neural Rendering with Per-View
  Optimization}.
\newblock \bibinfo{journal}{\emph{Computer Graphics Forum}}
  (\bibinfo{year}{2021}).
\newblock


\bibitem[Kuang et~al\mbox{.}(2022)]%
        {kuang2022neroic}
\bibfield{author}{\bibinfo{person}{Zhengfei Kuang}, \bibinfo{person}{Kyle
  Olszewski}, \bibinfo{person}{Menglei Chai}, \bibinfo{person}{Zeng Huang},
  \bibinfo{person}{Panos Achlioptas}, {and} \bibinfo{person}{Sergey Tulyakov}.}
  \bibinfo{year}{2022}\natexlab{}.
\newblock \showarticletitle{{NeROIC}: Neural Rendering of Objects from Online
  Image Collections}.
\newblock \bibinfo{journal}{\emph{SIGGRAPH}} (\bibinfo{year}{2022}).
\newblock


\bibitem[Labatut et~al\mbox{.}(2007)]%
        {labatut2007efficient}
\bibfield{author}{\bibinfo{person}{Patrick Labatut},
  \bibinfo{person}{Jean-Philippe Pons}, {and} \bibinfo{person}{Renaud
  Keriven}.} \bibinfo{year}{2007}\natexlab{}.
\newblock \showarticletitle{Efficient multi-view reconstruction of large-scale
  scenes using interest points, {Delaunay} triangulation and graph cuts}.
\newblock \bibinfo{journal}{\emph{ICCV}} (\bibinfo{year}{2007}).
\newblock


\bibitem[Levoy and Hanrahan(1996)]%
        {levoy1996light}
\bibfield{author}{\bibinfo{person}{Marc Levoy} {and} \bibinfo{person}{Pat
  Hanrahan}.} \bibinfo{year}{1996}\natexlab{}.
\newblock \showarticletitle{Light field rendering}.
\newblock \bibinfo{journal}{\emph{SIGGRAPH}} (\bibinfo{year}{1996}).
\newblock


\bibitem[Lombardi et~al\mbox{.}(2019)]%
        {lombardi2019neuralvolumes}
\bibfield{author}{\bibinfo{person}{Stephen Lombardi}, \bibinfo{person}{Tomas
  Simon}, \bibinfo{person}{Jason Saragih}, \bibinfo{person}{Gabriel Schwartz},
  \bibinfo{person}{Andreas Lehrmann}, {and} \bibinfo{person}{Yaser Sheikh}.}
  \bibinfo{year}{2019}\natexlab{}.
\newblock \showarticletitle{Neural Volumes: Learning Dynamic Renderable Volumes
  from Images}.
\newblock \bibinfo{journal}{\emph{SIGGRAPH}} (\bibinfo{year}{2019}).
\newblock


\bibitem[Lorensen and Cline(1987)]%
        {lorensen1987marching}
\bibfield{author}{\bibinfo{person}{W.~E. Lorensen} {and} \bibinfo{person}{H.~E.
  Cline}.} \bibinfo{year}{1987}\natexlab{}.
\newblock \showarticletitle{Marching cubes: A high resolution {3D} surface
  construction algorithm}.
\newblock \bibinfo{journal}{\emph{SIGGRAPH}} (\bibinfo{year}{1987}).
\newblock


\bibitem[Max(1995)]%
        {max1995optical}
\bibfield{author}{\bibinfo{person}{Nelson Max}.}
  \bibinfo{year}{1995}\natexlab{}.
\newblock \showarticletitle{Optical models for direct volume rendering}.
\newblock \bibinfo{journal}{\emph{IEEE TVCG}} (\bibinfo{year}{1995}).
\newblock


\bibitem[Mildenhall et~al\mbox{.}(2020)]%
        {mildenhall2020nerf}
\bibfield{author}{\bibinfo{person}{Ben Mildenhall}, \bibinfo{person}{Pratul~P.
  Srinivasan}, \bibinfo{person}{Matthew Tancik}, \bibinfo{person}{Jonathan~T.
  Barron}, \bibinfo{person}{Ravi Ramamoorthi}, {and} \bibinfo{person}{Ren Ng}.}
  \bibinfo{year}{2020}\natexlab{}.
\newblock \showarticletitle{{NeRF}: Representing Scenes as Neural Radiance
  Fields for View Synthesis}.
\newblock \bibinfo{journal}{\emph{ECCV}} (\bibinfo{year}{2020}).
\newblock


\bibitem[M{\"u}ller et~al\mbox{.}(2022)]%
        {muller2022instant}
\bibfield{author}{\bibinfo{person}{Thomas M{\"u}ller}, \bibinfo{person}{Alex
  Evans}, \bibinfo{person}{Christoph Schied}, {and} \bibinfo{person}{Alexander
  Keller}.} \bibinfo{year}{2022}\natexlab{}.
\newblock \showarticletitle{Instant neural graphics primitives with a
  multiresolution hash encoding}.
\newblock \bibinfo{journal}{\emph{SIGGRAPH}} (\bibinfo{year}{2022}).
\newblock


\bibitem[Neff et~al\mbox{.}(2021)]%
        {neff2021donerf}
\bibfield{author}{\bibinfo{person}{Thomas Neff}, \bibinfo{person}{Pascal
  Stadlbauer}, \bibinfo{person}{Mathias Parger}, \bibinfo{person}{Andreas
  Kurz}, \bibinfo{person}{Joerg~H. Mueller}, \bibinfo{person}{Chakravarty
  R.~Alla Chaitanya}, \bibinfo{person}{Anton Kaplanyan}, {and}
  \bibinfo{person}{Markus Steinberger}.} \bibinfo{year}{2021}\natexlab{}.
\newblock \showarticletitle{{DONeRF}: Towards Real-Time Rendering of Compact
  Neural Radiance Fields using Depth Oracle Networks}.
\newblock \bibinfo{journal}{\emph{Computer Graphics Forum}}
  (\bibinfo{year}{2021}).
\newblock


\bibitem[Oechsle et~al\mbox{.}(2021)]%
        {oechsle2021unisurf}
\bibfield{author}{\bibinfo{person}{Michael Oechsle}, \bibinfo{person}{Songyou
  Peng}, {and} \bibinfo{person}{Andreas Geiger}.}
  \bibinfo{year}{2021}\natexlab{}.
\newblock \showarticletitle{{UNISURF}: Unifying Neural Implicit Surfaces and
  Radiance Fields for Multi-View Reconstruction}.
\newblock \bibinfo{journal}{\emph{ICCV}} (\bibinfo{year}{2021}).
\newblock


\bibitem[Penner and Zhang(2017)]%
        {Soft3DReconstruction}
\bibfield{author}{\bibinfo{person}{Eric Penner} {and} \bibinfo{person}{Li
  Zhang}.} \bibinfo{year}{2017}\natexlab{}.
\newblock \showarticletitle{Soft 3D Reconstruction for View Synthesis}.
\newblock \bibinfo{journal}{\emph{SIGGRAPH Asia}} (\bibinfo{year}{2017}).
\newblock


\bibitem[Reiser et~al\mbox{.}(2021)]%
        {reiser2021kilonerf}
\bibfield{author}{\bibinfo{person}{Christian Reiser}, \bibinfo{person}{Songyou
  Peng}, \bibinfo{person}{Yiyi Liao}, {and} \bibinfo{person}{Andreas Geiger}.}
  \bibinfo{year}{2021}\natexlab{}.
\newblock \showarticletitle{{KiloNeRF}: Speeding up neural radiance fields with
  thousands of tiny {MLPs}}.
\newblock \bibinfo{journal}{\emph{ICCV}} (\bibinfo{year}{2021}).
\newblock


\bibitem[Riegler and Koltun(2020)]%
        {riegler2020free}
\bibfield{author}{\bibinfo{person}{Gernot Riegler} {and}
  \bibinfo{person}{Vladlen Koltun}.} \bibinfo{year}{2020}\natexlab{}.
\newblock \showarticletitle{Free View Synthesis}.
\newblock \bibinfo{journal}{\emph{ECCV}} (\bibinfo{year}{2020}).
\newblock


\bibitem[Riegler and Koltun(2021)]%
        {riegler2021stable}
\bibfield{author}{\bibinfo{person}{Gernot Riegler} {and}
  \bibinfo{person}{Vladlen Koltun}.} \bibinfo{year}{2021}\natexlab{}.
\newblock \showarticletitle{Stable view synthesis}.
\newblock \bibinfo{journal}{\emph{CVPR}} (\bibinfo{year}{2021}).
\newblock


\bibitem[R{\"u}ckert et~al\mbox{.}(2022)]%
        {ruckert2021adop}
\bibfield{author}{\bibinfo{person}{Darius R{\"u}ckert}, \bibinfo{person}{Linus
  Franke}, {and} \bibinfo{person}{Marc Stamminger}.}
  \bibinfo{year}{2022}\natexlab{}.
\newblock \showarticletitle{{ADOP}: Approximate differentiable one-pixel point
  rendering}.
\newblock \bibinfo{journal}{\emph{SIGGRAPH}} (\bibinfo{year}{2022}).
\newblock


\bibitem[Sander et~al\mbox{.}(2007)]%
        {Sander2007}
\bibfield{author}{\bibinfo{person}{Pedro~V. Sander}, \bibinfo{person}{Diego
  Nehab}, {and} \bibinfo{person}{Joshua Barczak}.}
  \bibinfo{year}{2007}\natexlab{}.
\newblock \showarticletitle{Fast Triangle Reordering for Vertex Locality and
  Reduced Overdraw}.
\newblock \bibinfo{journal}{\emph{SIGGRAPH}} (\bibinfo{year}{2007}).
\newblock


\bibitem[Sch\"{o}nberger et~al\mbox{.}(2016)]%
        {schoenberger2016mvs}
\bibfield{author}{\bibinfo{person}{Johannes~Lutz Sch\"{o}nberger},
  \bibinfo{person}{Enliang Zheng}, \bibinfo{person}{Marc Pollefeys}, {and}
  \bibinfo{person}{Jan-Michael Frahm}.} \bibinfo{year}{2016}\natexlab{}.
\newblock \showarticletitle{Pixelwise View Selection for Unstructured
  Multi-View Stereo}.
\newblock \bibinfo{journal}{\emph{ECCV}} (\bibinfo{year}{2016}).
\newblock


\bibitem[Srinivasan et~al\mbox{.}(2021)]%
        {srinivasan2021nerv}
\bibfield{author}{\bibinfo{person}{Pratul~P. Srinivasan},
  \bibinfo{person}{Boyang Deng}, \bibinfo{person}{Xiuming Zhang},
  \bibinfo{person}{Matthew Tancik}, \bibinfo{person}{Ben Mildenhall}, {and}
  \bibinfo{person}{Jonathan~T. Barron}.} \bibinfo{year}{2021}\natexlab{}.
\newblock \showarticletitle{{NeRV}: Neural reflectance and visibility fields
  for relighting and view synthesis}.
\newblock \bibinfo{journal}{\emph{CVPR}} (\bibinfo{year}{2021}).
\newblock


\bibitem[Srinivasan et~al\mbox{.}(2019)]%
        {srinivasan2019pushing}
\bibfield{author}{\bibinfo{person}{Pratul~P. Srinivasan},
  \bibinfo{person}{Richard Tucker~Joand nathan T.~Barron},
  \bibinfo{person}{Ravi Ramamoorthi}, \bibinfo{person}{Ren Ng}, {and}
  \bibinfo{person}{Noah Snavely}.} \bibinfo{year}{2019}\natexlab{}.
\newblock \showarticletitle{Pushing the Boundaries of View Extrapolation with
  Multiplane Images}.
\newblock \bibinfo{journal}{\emph{CVPR}} (\bibinfo{year}{2019}).
\newblock


\bibitem[Sun et~al\mbox{.}(2022)]%
        {sun2022direct}
\bibfield{author}{\bibinfo{person}{Cheng Sun}, \bibinfo{person}{Min Sun}, {and}
  \bibinfo{person}{Hwann-Tzong Chen}.} \bibinfo{year}{2022}\natexlab{}.
\newblock \showarticletitle{Direct voxel grid optimization: Super-fast
  convergence for radiance fields reconstruction}.
\newblock \bibinfo{journal}{\emph{CVPR}} (\bibinfo{year}{2022}).
\newblock


\bibitem[Szeliski and Golland(1999)]%
        {szeliski1999planesweep}
\bibfield{author}{\bibinfo{person}{Richard Szeliski} {and}
  \bibinfo{person}{Polina Golland}.} \bibinfo{year}{1999}\natexlab{}.
\newblock \showarticletitle{Stereo Matching with Transparency and Matting}.
\newblock \bibinfo{journal}{\emph{IJCV}} (\bibinfo{year}{1999}).
\newblock


\bibitem[Tewari et~al\mbox{.}(2022)]%
        {tewari2022advances}
\bibfield{author}{\bibinfo{person}{Ayush Tewari}, \bibinfo{person}{Justus
  Thies}, \bibinfo{person}{Ben Mildenhall}, \bibinfo{person}{Pratul
  Srinivasan}, \bibinfo{person}{Edgar Tretschk}, \bibinfo{person}{W Yifan},
  \bibinfo{person}{Christoph Lassner}, \bibinfo{person}{Vincent Sitzmann},
  \bibinfo{person}{Ricardo Martin-Brualla}, \bibinfo{person}{Stephen Lombardi},
  {et~al\mbox{.}}} \bibinfo{year}{2022}\natexlab{}.
\newblock \showarticletitle{Advances in neural rendering}.
\newblock \bibinfo{journal}{\emph{Computer Graphics Forum}}
  (\bibinfo{year}{2022}).
\newblock


\bibitem[Verbin et~al\mbox{.}(2022)]%
        {verbin2022ref}
\bibfield{author}{\bibinfo{person}{Dor Verbin}, \bibinfo{person}{Peter Hedman},
  \bibinfo{person}{Ben Mildenhall}, \bibinfo{person}{Todd Zickler},
  \bibinfo{person}{Jonathan~T Barron}, {and} \bibinfo{person}{Pratul~P
  Srinivasan}.} \bibinfo{year}{2022}\natexlab{}.
\newblock \showarticletitle{{Ref-NeRF}: Structured view-dependent appearance
  for neural radiance fields}.
\newblock \bibinfo{journal}{\emph{CVPR}} (\bibinfo{year}{2022}).
\newblock


\bibitem[Vogiatzis et~al\mbox{.}(2007)]%
        {vogiatzis2007volumetric}
\bibfield{author}{\bibinfo{person}{G. Vogiatzis}, \bibinfo{person}{C.
  Hern\'andez}, \bibinfo{person}{P. Torr}, {and} \bibinfo{person}{R. Cipolla}.}
  \bibinfo{year}{2007}\natexlab{}.
\newblock \showarticletitle{Multi-View Stereo via Volumetric Graph-Cuts and
  Occlusion Robust Photo-Consistency}.
\newblock \bibinfo{journal}{\emph{IEEE TPAMI}} (\bibinfo{year}{2007}).
\newblock


\bibitem[Wang et~al\mbox{.}(2021)]%
        {wang2021neus}
\bibfield{author}{\bibinfo{person}{Peng Wang}, \bibinfo{person}{Lingjie Liu},
  \bibinfo{person}{Yuan Liu}, \bibinfo{person}{Christian Theobalt},
  \bibinfo{person}{Taku Komura}, {and} \bibinfo{person}{Wenping Wang}.}
  \bibinfo{year}{2021}\natexlab{}.
\newblock \showarticletitle{{NeuS}: Learning Neural Implicit Surfaces by Volume
  Rendering for Multi-view Reconstruction}.
\newblock \bibinfo{journal}{\emph{NeurIPS}} (\bibinfo{year}{2021}).
\newblock


\bibitem[Wizadwongsa et~al\mbox{.}(2021)]%
        {wizadwongsa2021nex}
\bibfield{author}{\bibinfo{person}{Suttisak Wizadwongsa},
  \bibinfo{person}{Pakkapon Phongthawee}, \bibinfo{person}{Jiraphon
  Yenphraphai}, {and} \bibinfo{person}{Supasorn Suwajanakorn}.}
  \bibinfo{year}{2021}\natexlab{}.
\newblock \showarticletitle{{NeX}: Real-time View Synthesis with Neural Basis
  Expansion}.
\newblock \bibinfo{journal}{\emph{CVPR}} (\bibinfo{year}{2021}).
\newblock


\bibitem[Wood et~al\mbox{.}(2000)]%
        {wood:2000:slf}
\bibfield{author}{\bibinfo{person}{Daniel~N. Wood}, \bibinfo{person}{Daniel~I.
  Azuma}, \bibinfo{person}{Ken Aldinger}, \bibinfo{person}{Brian Curless},
  \bibinfo{person}{Tom Duchamp}, \bibinfo{person}{David~H. Salesin}, {and}
  \bibinfo{person}{Werner Stuetzle}.} \bibinfo{year}{2000}\natexlab{}.
\newblock \showarticletitle{Surface Light Fields for {3D} Photography}.
\newblock \bibinfo{journal}{\emph{SIGGRAPH}} (\bibinfo{year}{2000}).
\newblock


\bibitem[Wu et~al\mbox{.}(2022)]%
        {wu2022snisr}
\bibfield{author}{\bibinfo{person}{Xiuchao Wu}, \bibinfo{person}{Jiamin Xu},
  \bibinfo{person}{Zihan Zhu}, \bibinfo{person}{Hujun Bao},
  \bibinfo{person}{Qixing Huang}, \bibinfo{person}{James Tompkin}, {and}
  \bibinfo{person}{Weiwei Xu}.} \bibinfo{year}{2022}\natexlab{}.
\newblock \showarticletitle{Scalable Neural Indoor Scene Rendering}.
\newblock \bibinfo{journal}{\emph{ACM TOG}} (\bibinfo{year}{2022}).
\newblock


\bibitem[Yariv et~al\mbox{.}(2021)]%
        {yariv2021volume}
\bibfield{author}{\bibinfo{person}{Lior Yariv}, \bibinfo{person}{Jiatao Gu},
  \bibinfo{person}{Yoni Kasten}, {and} \bibinfo{person}{Yaron Lipman}.}
  \bibinfo{year}{2021}\natexlab{}.
\newblock \showarticletitle{Volume rendering of neural implicit surfaces}.
\newblock \bibinfo{journal}{\emph{NeurIPS}} (\bibinfo{year}{2021}).
\newblock


\bibitem[Yu et~al\mbox{.}(2022)]%
        {yu2021plenoxels}
\bibfield{author}{\bibinfo{person}{Alex Yu}, \bibinfo{person}{Sara
  Fridovich-Keil}, \bibinfo{person}{Matthew Tancik}, \bibinfo{person}{Qinhong
  Chen}, \bibinfo{person}{Benjamin Recht}, {and} \bibinfo{person}{Angjoo
  Kanazawa}.} \bibinfo{year}{2022}\natexlab{}.
\newblock \showarticletitle{Plenoxels: Radiance fields without neural
  networks}.
\newblock \bibinfo{journal}{\emph{CVPR}} (\bibinfo{year}{2022}).
\newblock


\bibitem[Yu et~al\mbox{.}(2021)]%
        {yu2021plenoctrees}
\bibfield{author}{\bibinfo{person}{Alex Yu}, \bibinfo{person}{Ruilong Li},
  \bibinfo{person}{Matthew Tancik}, \bibinfo{person}{Hao Li},
  \bibinfo{person}{Ren Ng}, {and} \bibinfo{person}{Angjoo Kanazawa}.}
  \bibinfo{year}{2021}\natexlab{}.
\newblock \showarticletitle{{PlenOctrees} for real-time rendering of neural
  radiance fields}.
\newblock \bibinfo{journal}{\emph{ICCV}} (\bibinfo{year}{2021}).
\newblock


\bibitem[Zhang et~al\mbox{.}(2021a)]%
        {zhang2021physg}
\bibfield{author}{\bibinfo{person}{Kai Zhang}, \bibinfo{person}{Fujun Luan},
  \bibinfo{person}{Qianqian Wang}, \bibinfo{person}{Kavita Bala}, {and}
  \bibinfo{person}{Noah Snavely}.} \bibinfo{year}{2021}\natexlab{a}.
\newblock \showarticletitle{{PhySG}: Inverse Rendering with Spherical Gaussians
  for Physics-based Material Editing and Relighting}.
\newblock \bibinfo{journal}{\emph{CVPR}} (\bibinfo{year}{2021}).
\newblock


\bibitem[Zhang et~al\mbox{.}(2020)]%
        {kaizhang2020}
\bibfield{author}{\bibinfo{person}{Kai Zhang}, \bibinfo{person}{Gernot
  Riegler}, \bibinfo{person}{Noah Snavely}, {and} \bibinfo{person}{Vladlen
  Koltun}.} \bibinfo{year}{2020}\natexlab{}.
\newblock \showarticletitle{{NeRF++}: Analyzing and Improving Neural Radiance
  Fields}.
\newblock \bibinfo{journal}{\emph{arXiv:2010.07492}} (\bibinfo{year}{2020}).
\newblock


\bibitem[Zhang et~al\mbox{.}(2021b)]%
        {zhang2021nerfactor}
\bibfield{author}{\bibinfo{person}{Xiuming Zhang}, \bibinfo{person}{Pratul~P.
  Srinivasan}, \bibinfo{person}{Boyang Deng}, \bibinfo{person}{Paul Debevec},
  \bibinfo{person}{William~T. Freeman}, {and} \bibinfo{person}{Jonathan~T.
  Barron}.} \bibinfo{year}{2021}\natexlab{b}.
\newblock \showarticletitle{{NeRFactor}: Neural Factorization of Shape and
  Reflectance Under an Unknown Illumination}.
\newblock \bibinfo{journal}{\emph{SIGGRAPH Asia}} (\bibinfo{year}{2021}).
\newblock


\bibitem[Zhou et~al\mbox{.}(2018)]%
        {Zhou_StereoMagn_SG18}
\bibfield{author}{\bibinfo{person}{Tinghui Zhou}, \bibinfo{person}{Richard
  Tucker}, \bibinfo{person}{John Flynn}, \bibinfo{person}{Graham Fyffe}, {and}
  \bibinfo{person}{Noah Snavely}.} \bibinfo{year}{2018}\natexlab{}.
\newblock \showarticletitle{Stereo Magnification: Learning View Synthesis Using
  Multiplane Images}.
\newblock \bibinfo{journal}{\emph{SIGGRAPH}} (\bibinfo{year}{2018}).
\newblock


\end{thebibliography}

\newpage
\appendix

\section{Training and optimization details}
\paragraph{SDF model definition and optimization.}
As stated in Section 4.1, we model our SDF using a variant of mip-NeRF 360.
We train our model
using the same optimization settings as mip-NeRF 360 (250k iterations of Adam~\cite{adam} with a batch size of $2^{14}$ and a learning rate that is warm-started and then log-linearly interpolated from $2\cdot 10^{-3}$ to $2\cdot 10^{-5}$,
with $\beta_1 = 0.9$, $\beta_2=0.999$, $\epsilon=10^{-6}$) and similar MLP architectures (a proposal MLP with 4 layers and 256 hidden units, and a NeRF MLP with 8 layers and 1024 hidden units, both using swish/SiLU rectifiers~\cite{hendrycks2016gaussian} and 8 scales of positional encoding).
Following the hierarchical sampling procedure of mip-NeRF 360, we perform two resampling stages using 64 samples evaluated using the proposal MLP, and then one evaluation stage using 32 samples of the NeRF MLP. The proposal MLP is optimized by minimizing $\mathcal{L_{\mathrm{prop}}} + 0.1 \mathcal{L_{\mathrm{SDF}}}$ where $\mathcal{L_{\mathrm{prop}}}$ is the proposal loss described in~\cite{barron2022mipnerf360}, designed to bound the weights output by the NeRF MLP density. 

\paragraph{Optimizing for per-vertex attributes via a compressed hash grid.} 
As stated Section 4.3, during optimization we use Instant NGP~\cite{muller2022instant} as the underlying representation for our vertex attributes. We use the following hyperparameters: $\textrm{L}=18$, $\textrm{T}=2^{21}$ and $\textrm{N}_\textrm{max}=8192$. We remove the view-direction input from the NGP model, as we incorporate it later in Equation~7. We use a weight decay of $0.1$ for the hash grids but not the MLP, optimize using Adam~\cite{adam} for 150k iterations with a batch size of $2^{14}$ and an initial learning rate of $0.001$ that we drop by $10\times$ every 50k iterations.

\section{Tweaks for a compelling viewer}
Here we detail a few tweaks to the pipeline which do not strictly improve reconstruction accuracy, but rather makes for a more compelling viewing experience. With this in mind, we found it important to alleviate jarring transitions between the reconstructed scene content and the background color. To this end, we also include a global clear color into the appearance parameters we optimize for in Section 4.3. That is, we assign this color to any pixel in the training data which does not have a valid triangle index. 

To further mask the transition between geometry and background, we enclose SDF with bounding geometry before extracting the mesh in Section 4.2.
We compute a convex hull computed as the intersection of $32$ randomly oriented planes, where the location of each plane has been set to bound $99.75\%$ of the voxels that have marked as candidates for surface extraction. We then further make this hull conservative by inflating it by a slight margin of $\times1.025$. 
However, since the extracted mesh needs to be transformed into world space for rendering, we must take care to avoid numerical precision issues that may arise from using unbounded vertex coordinates during rasterization. We solve this by bounding the scene with a distant sphere with a radius of $500$ world-space units. These two operations are easily implemented by setting the SDF value in each grid cell to the pointwise minimum of the MLP-parameterized SDF and the SDF of the defined bounding geometry.

\section{Baselines details}
\paragraph{MobileNeRF viewer configuration} 
Note that by default the MobileNeRF viewer runs at a reduced resolution for high-framerates across a variety of devices. For our comparisons we modify it to run at different resolutions. When we compute image quality metrics, we choose the resolution of the test set images. Furthermore, when we measure run-time performance we use a $1920\times1080$, which is a resolution that is representative for most modern displays.

\paragraph{Instant NGP} Table~\ref{tab:alldoor} reports quality results for Instant NGP~\cite{muller2022instant} method, where we carefully adapt it to work on unbounded large scenes. We asked the authors of Instant NGP for help with tuning their method and made the following changes:
\begin{itemize}
  \item We use \textit{big.json} configuration file provided with the official code release,
  \item we increased the batch size by $4\times$ to $2^{20}$, and
  \item we increased the scene scale from $16$ to $32$.
\end{itemize}
Note that none of these changes has a significant impact on the render time for Instant NGP.

By default, the Instant NGP viewer is equipped with a dynamic upscaling implementation, which renders images at a lower resolution and then applies smart upscaling. For a fair comparison we turn this off when measuring perfomance, as these dynamic upscalers can be applied to any renderer. More importantly, we want the perfomance numbers to correspond with the test set quality metrics, and none of the test-set images were computed using upscaling.

\end{document}